\documentclass[10pt,journal,compsoc]{IEEEtran}

\ifCLASSOPTIONcompsoc
  \usepackage[nocompress]{cite}
  \usepackage[caption=false,font=footnotesize,labelfont=sf,textfont=sf]{subfig}
\else
  \usepackage{cite}
  \usepackage[caption=false,font=footnotesize]{subfig}
\fi
\ifCLASSINFOpdf
  \usepackage[pdftex]{graphicx}
  \DeclareGraphicsExtensions{.pdf,.jpeg,.png}
\else
  \usepackage[dvips]{graphicx}
  \DeclareGraphicsExtensions{.eps}
\fi
\usepackage{amsmath,amssymb,amsfonts,dsfont,bm,mathrsfs}
\usepackage{algorithmic}
\usepackage{booktabs,multirow,adjustbox}
\usepackage{float}
\usepackage[table]{xcolor}
\usepackage{collcell}
\usepackage{xifthen}
\usepackage{wrapfig}
\usepackage{bbding}
\usepackage{pgf}
\usepackage[pagebackref=false,breaklinks=true,colorlinks=true,citecolor=blue,bookmarks=false]{hyperref}
\def\ColorMode{hsb}
\newcommand{\ColCell}[1]{
  \ifthenelse{\isempty{#1}}{}{
    \pgfmathparse{#1<2?1:0}                          
      \ifnum\pgfmathresult=0\relax\color{white}\fi
    \pgfmathsetmacro\compA{0}                        
    \pgfmathsetmacro\compB{(abs(#1)==1?0:abs(#1))/1} 
    \pgfmathsetmacro\compC{1}                        
    \edef\x{\noexpand\centering\noexpand\cellcolor[\ColorMode]{\compA,\compB,\compC}}\x #1
  }
}
\newcolumntype{C}[1]{>{\collectcell\ColCell}m{#1}<{\endcollectcell}}  

\newcommand{\R}{\mathbb{R}}      
\newcommand{\E}{\mathbb{E}}      
\newcommand{\0}{{\rm\bf 0}}      
\newcommand{\I}{{\rm\bf I}}      
\newcommand{\z}{{\rm\bf z}}      
\newcommand{\Z}{\mathcal{Z}}     
\newcommand{\w}{{\rm\bf w}}      
\newcommand{\W}{\mathcal{W}}     
\newcommand{\x}{{\rm\bf x}}      
\newcommand{\X}{\mathcal{X}}     
\newcommand{\s}{{\rm\bf s}}      
\renewcommand{\S}{\mathcal{S}}   
\newcommand{\n}{{\rm\bf n}}      
\newcommand{\N}{{\rm\bf N}}      

\begin{document}

\title{Zero-Shot Object Counting \\ with Language-Vision Models}

\def\mA{\mathcal{A}}
\def\mB{\mathcal{B}}
\def\mC{\mathcal{C}}
\def\mD{\mathcal{D}}
\def\mE{\mathcal{E}}
\def\mF{\mathcal{F}}
\def\mG{\mathcal{G}}
\def\mH{\mathcal{H}}
\def\mI{\mathcal{I}}
\def\mJ{\mathcal{J}}
\def\mK{\mathcal{K}}
\def\mL{\mathcal{L}}
\def\mM{\mathcal{M}}
\def\mN{\mathcal{N}}
\def\mO{\mathcal{O}}
\def\mP{\mathcal{P}}
\def\mQ{\mathcal{Q}}
\def\mR{\mathcal{R}}
\def\mS{\mathcal{S}}
\def\mT{\mathcal{T}}
\def\mU{\mathcal{U}}
\def\mV{\mathcal{V}}
\def\mW{\mathcal{W}}
\def\mX{\mathcal{X}}
\def\mY{\mathcal{Y}}
\def\mZ{\mathcal{Z}}

\def\1n{\mathbf{1}_n}
\def\0{\mathbf{0}}
\def\1{\mathbf{1}}

\def\A{{\bf A}}
\def\B{{\bf B}}
\def\C{{\bf C}}
\def\D{{\bf D}}
\def\E{{\bf E}}
\def\F{{\bf F}}
\def\G{{\bf G}}
\def\H{{\bf H}}
\def\I{{\bf I}}
\def\J{{\bf J}}
\def\K{{\bf K}}
\def\L{{\bf L}}
\def\M{{\bf M}}
\def\N{{\bf N}}
\def\O{{\bf O}}
\def\P{{\bf P}}
\def\Q{{\bf Q}}
\def\R{{\bf R}}
\def\S{{\bf S}}
\def\T{{\bf T}}
\def\U{{\bf U}}
\def\V{{\bf V}}
\def\W{{\bf W}}
\def\X{{\bf X}}
\def\Y{{\bf Y}}
\def\Z{{\bf Z}}

\def\a{{\bf a}}
\def\b{{\bf b}}
\def\c{{\bf c}}
\def\d{{\bf d}}
\def\e{{\bf e}}
\def\f{{\bf f}}
\def\g{{\bf g}}
\def\h{{\bf h}}
\def\i{{\bf i}}
\def\j{{\bf j}}
\def\k{{\bf k}}
\def\l{{\bf l}}
\def\m{{\bf m}}
\def\n{{\bf n}}
\def\o{{\bf o}}
\def\p{{\bf p}}
\def\q{{\bf q}}
\def\r{{\bf r}}
\def\s{{\bf s}}
\def\t{{\bf t}}
\def\u{{\bf u}}
\def\v{{\bf v}}
\def\w{{\bf w}}
\def\x{{\bf x}}
\def\y{{\bf y}}
\def\z{{\bf z}}

\def\balpha{\mbox{\boldmath{$\alpha$}}}
\def\bbeta{\mbox{\boldmath{$\beta$}}}
\def\bdelta{\mbox{\boldmath{$\delta$}}}
\def\bgamma{\mbox{\boldmath{$\gamma$}}}
\def\blambda{\mbox{\boldmath{$\lambda$}}}
\def\bsigma{\mbox{\boldmath{$\sigma$}}}
\def\btheta{\mbox{\boldmath{$\theta$}}}
\def\bomega{\mbox{\boldmath{$\omega$}}}
\def\bxi{\mbox{\boldmath{$\xi$}}}
\def\bnu{\mbox{\boldmath{$\nu$}}}                                  
\def\bphi{\mbox{\boldmath{$\phi$}}}

\def\bDelta{\mbox{\boldmath{$\Delta$}}}
\def\bOmega{\mbox{\boldmath{$\Omega$}}}
\def\bPhi{\mbox{\boldmath{$\Phi$}}}
\def\bLambda{\mbox{\boldmath{$\Lambda$}}}
\def\bSigma{\mbox{\boldmath{$\Sigma$}}}
\def\bGamma{\mbox{\boldmath{$\Gamma$}}}

\newcommand{\myminimum}[1]{\mathop{\textrm{minimum}}_{#1}}
\newcommand{\mymaximum}[1]{\mathop{\textrm{maximum}}_{#1}}    
\newcommand{\mymean}[1]{\mathop{\textrm{mean}}_{#1}}
\newcommand{\myvar}[1]{\mathop{\textrm{Variance}}_{#1}}
\newcommand{\mymin}[1]{\mathop{\textrm{minimize}}_{#1}}
\newcommand{\mymax}[1]{\mathop{\textrm{maximize}}_{#1}}
\newcommand{\mymins}[1]{\mathop{\textrm{min.}}_{#1}}
\newcommand{\mymaxs}[1]{\mathop{\textrm{max.}}_{#1}}  
\newcommand{\myargmin}[1]{\mathop{\textrm{argmin}}_{#1}} 
\newcommand{\myargmax}[1]{\mathop{\textrm{argmax}}_{#1}} 
\newcommand{\myst}{\textrm{s.t. }}

\newcommand{\denselist}{\itemsep -1pt}
\newcommand{\sparselist}{\itemsep 1pt}

\newcommand{\cyan}[1]{\textcolor{cyan}{#1}}
\newcommand{\red}[1]{\textcolor{red}{#1}}  
\newcommand{\blue}[1]{\textcolor{blue}{#1}}
\newcommand{\magenta}[1]{\textcolor{magenta}{#1}}
\newcommand{\pink}[1]{\textcolor{pink}{#1}}
\newcommand{\green}[1]{\textcolor{green}{#1}} 
\newcommand{\gray}[1]{\textcolor{gray}{#1}}    
\newcommand{\mygreen}[1]{\textcolor{mygreen}{#1}}    
\newcommand{\purple}[1]{\textcolor{purple}{#1}}

\newcommand{\mtodo}[1]{{\color{red}$\blacksquare$\textbf{[TODO: #1]}}}
\newcommand{\myheading}[1]{\vspace{1ex}\noindent \textbf{#1}}

\def\changemargin#1#2{\list{}{\rightmargin#2\leftmargin#1}\item[]}
\let\endchangemargin=\endlist
                                               
\newcommand{\cm}[1]{}

\def\xbi{\overline{\x}_i}
\def\wbi{\overline{\w}_{(i)}}
\def\wb{\overline{\w}}
\def\Ib{\overline{\I}}
\def\invC{\C^{-1}}
\def\invCi{\C_{(i)}^{-1}}
\def\ab{\overline{\balpha}}
\def\abi{\overline{\balpha}_{(i)}}
\def\Kb{\overline{\K}}
\def\Xb{\overline{\X}}
\def\kbi{\overline{\k}_{i}}
\def\Kzz{\K_{\z\z}}
\def\Kzx{\K_{\z\x}}
\def\Xsub{\X_{sub}}
\def\ssub{\s_{sub}}
\def\wbsub{\overline{\w}_{sub}}
\def\dsub{\d_{sub}}
\def\invCsub{\C^{-1}_{sub}}
\def\etal{\emph{et al}.}
\def\etals{\emph{et al}. }
\def\DS{\textcolor{red}}
\newcommand{\norm}[1]{\left\lVert#1\right\rVert}
\def\subFigSzab{\linewidth}
%
%
%
%

\author{Jingyi Xu, Hieu Le and~Dimitris Samaras 
 \IEEEcompsocitemizethanks{\IEEEcompsocthanksitem{J. Xu and D. Samaras are with the Department of Computer Science, Stony Brook University, Stony Brook, NY 11794. E-mail: \{jingyixu, samaras\}@cs.stonybrook.edu}\IEEEcompsocthanksitem{H. Le is with the Computer Vision Lab, EPFL, Laussane, Switzerland, E-mail: minh.le@epfl.ch} }
}



%
%

\markboth{IEEE TRANSACTIONS ON PATTERN ANALYSIS AND MACHINE INTELLIGENCE }%
{Shell \MakeLowercase{\textit{et al.}}: Bare Demo of IEEEtran.cls for Computer Society Journals}
%



\IEEEtitleabstractindextext{%
\begin{abstract}
Class-agnostic object counting aims to count object instances of an arbitrary class at test time. It is challenging but also enables many potential applications. Current methods require human-annotated exemplars as inputs which are often unavailable for novel categories, especially for autonomous systems. Thus, we propose zero-shot object counting (ZSC), a new setting where only the class name is available during test time. This obviates the need for human annotators and enables automated operation. 
To perform ZSC, we propose finding a few object crops from the input image and use them as counting exemplars.  The goal is to identify patches containing the objects of interest while also being visually representative for all instances in the image. To do this, we first construct class prototypes using large language-vision models, including CLIP and Stable Diffusion, to select the patches containing the target objects. Furthermore, we propose a ranking model that estimates the counting error of each patch to select the most suitable exemplars for counting.  Experimental results on a recent class-agnostic counting dataset, FSC-147, validate the effectiveness of our method.
\end{abstract}




}

\maketitle
 \IEEEdisplaynontitleabstractindextext
\IEEEpeerreviewmaketitle

\begin{figure}[t]
\begin{center}
\includegraphics[width=0.7\linewidth]{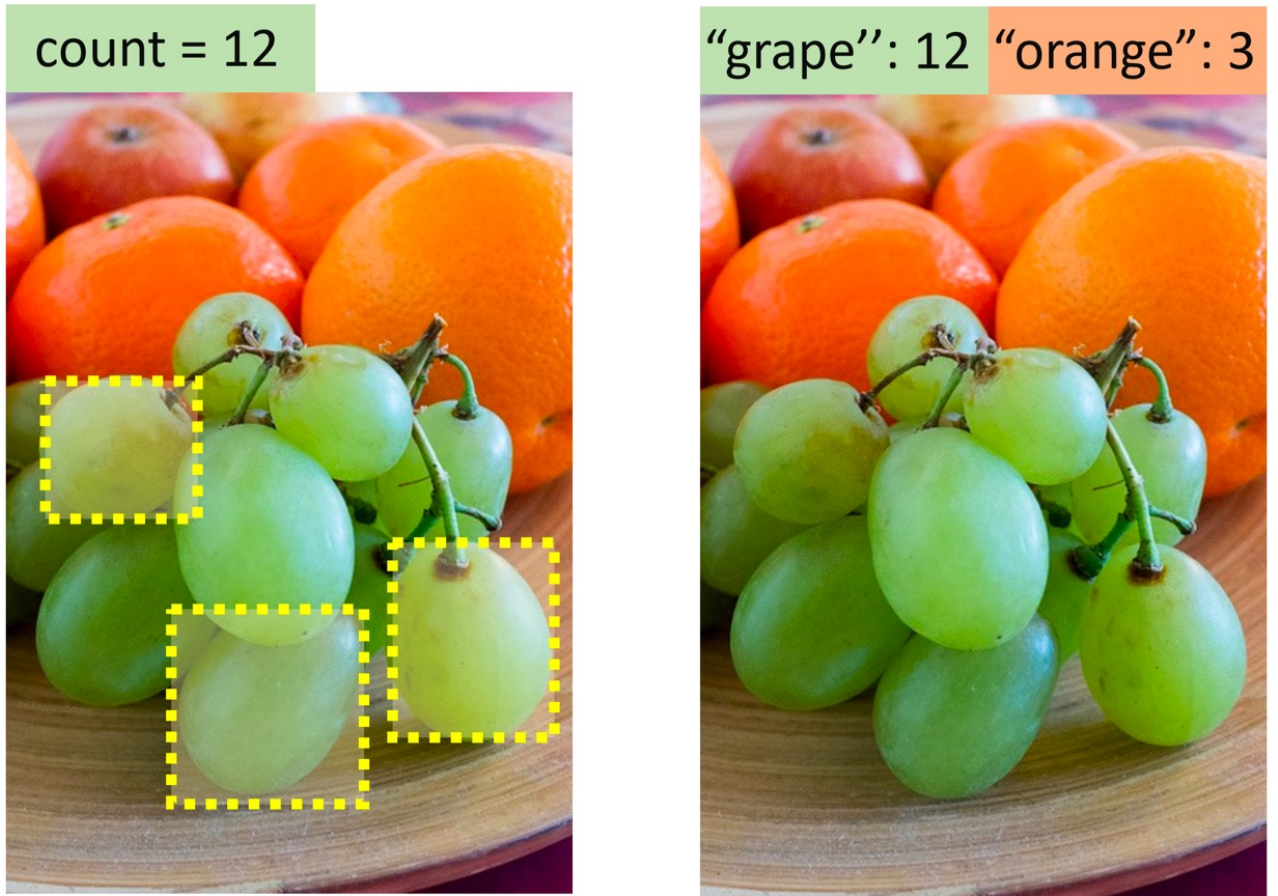}
\makebox[0.45\linewidth]{(a) Few-shot Counting }
\makebox[0.45\linewidth]{(b) Zero-Shot Counting}
\end{center} \vspace{2pt}
\caption{
Our proposed task of zero-shot object counting (ZSC). Traditional few-shot counting methods require a few exemplars of the object category (a). We propose zero-shot counting where the counter only needs the class name to count the number of object instances. (b). Few-shot counting methods require human annotators at test time while zero-shot counters can be fully automatic.
}  \vspace{-2pt}
\label{fig:teaser} 
\end{figure}

\IEEEraisesectionheading{\section{Introduction}\label{sec:introduction}}
Object counting aims to infer the number of objects in an image. Most of the existing methods focus on counting objects from specialized categories such as human crowds \cite{Sam2022SSCrowd}, cars \cite{Mundhenk2016ALC}, animals \cite{Arteta2016CountingIT}, and cells \cite{Xie2018MicroscopyCC}. These methods count only a single category at a time. 
Recently, class-agnostic counting \cite{Ranjan2021LearningTC,Shi2022SimiCounting,Lu2018CAC} has been proposed to count objects of arbitrary categories. Several human-annotated bounding boxes of objects are required to specify the objects of interest (see Figure \ref{fig:teaser}a). However, having humans in the loop is not practical for many real-world applications, such as fully automated wildlife monitoring systems or visual anomaly detection systems. 
%

A more practical setting, exemplar-free class-agnostic counting, has been proposed recently by Ranjan \textit{et al.}\cite{Ranjan2022Exemplar}. They introduce RepRPN, which first identifies the objects that occur most frequently in the image, and then uses them as exemplars for object counting. Even though RepRPN does not require any annotated boxes at test time, the method simply counts objects from the class with the highest number of instances. As a result, it can not be used for counting a specific class of interest. The method is only suitable for counting images with a single dominant object class, which limits its potential applicability.
 
Thus, our goal is to build an exemplar-free object counter where we can specify what to count.
To this end, we introduce a new counting task in which the user only needs to provide the name of the class for counting rather than the exemplars (see Figure \ref{fig:teaser}b). 
Note that the class to count during test time can be arbitrary. For cases where the test class is completely unseen to the trained model, the counter needs to adapt to the unseen class without any annotated data. Hence, we name this setting zero-shot object counting (ZSC), inspired by previous zero-shot learning approaches \cite{Zheng2021ZeroShotIS,Bansal2018ZeroShotOD}. 

To count without any annotated exemplars, we propose finding a few patches in the input image containing the target object to use them as counting exemplars. There are two challenges: 1) how to localize patches that contain the object of interest based on the provided class name, and 2) how to select \textit{good} exemplars for counting. Ideally, good object exemplars are visually representative for most instances in the image, which can benefit the object counter. In addition, we want to avoid selecting patches that contain irrelevant objects or backgrounds, which likely lead to incorrect object counts.  To this end, we propose a two-step method that first localizes the class-relevant patches which contain the objects of interest based on the given class name, and then selects among these patches the optimal exemplars for counting. We use these selected exemplars,  together with a pre-trained exemplar-based counting model, to achieve exemplar-free object counting.

The first step of our framework involves constructing a class prototype based on the given class name. Essentially, this requires a mapping between the categorical label and its visual feature. We employ pre-trained large language-vision models to accomplish this via two approaches.

In the conference version of this paper \cite{Xu2023zsc}, we learn this mapping between language queries and visual features via a conditional variational autoencoder (VAE). This VAE model is trained to generate visual features of object crops for any arbitrary class, conditioned on its semantic embedding extracted from a pre-trained language-vision model \cite{Radford2021LearningTV}. We take the average of the generated features to compute the class prototype, which can be then used to select class-relevant patches through a simple nearest-neighbour lookup scheme. In essence, our VAE-based approach creates a single prototypical feature for each category that can be applied to any images of this class. 



However, a single prototypical feature might not work well for categories with significant intra-class variance. Objects of the same category across different images can exhibit significant differences in colors (e.g., a green apple versus a red apple), shapes (e.g., an SUV versus a regular car), scales, or materials (a wooden versus a fabric chair). To better handle this variability, we propose to construct a class prototype specific to each image. To do so, we leverage the recent advancements in text-to-image generative models, i.e., Stable Diffusion \cite{Rombach2022LatentDiff}, for prototype generation. Compared to classic VAE-based models, Stable Diffusion enables more realistic and diverse sample generation thanks to large-scale training data. This provides a way to deal with the variations in query objects. Specifically, given a query image, we first use Stable Diffusion to generate a variety of images containing the objects of interest.
Then we select among them the object crops that most resemble the query objects and only use them for constructing the class prototype. 
In this way, a unique prototype is constructed specifically for each testing sample, as opposed to the VAE-based approach that uses a single universal prototype for all images. We show that using image-specific prototypes generally leads to better counting performance, compared to using a single generic categorical one.
%

\begin{figure}[t]
\begin{center}
\includegraphics[width=\linewidth]{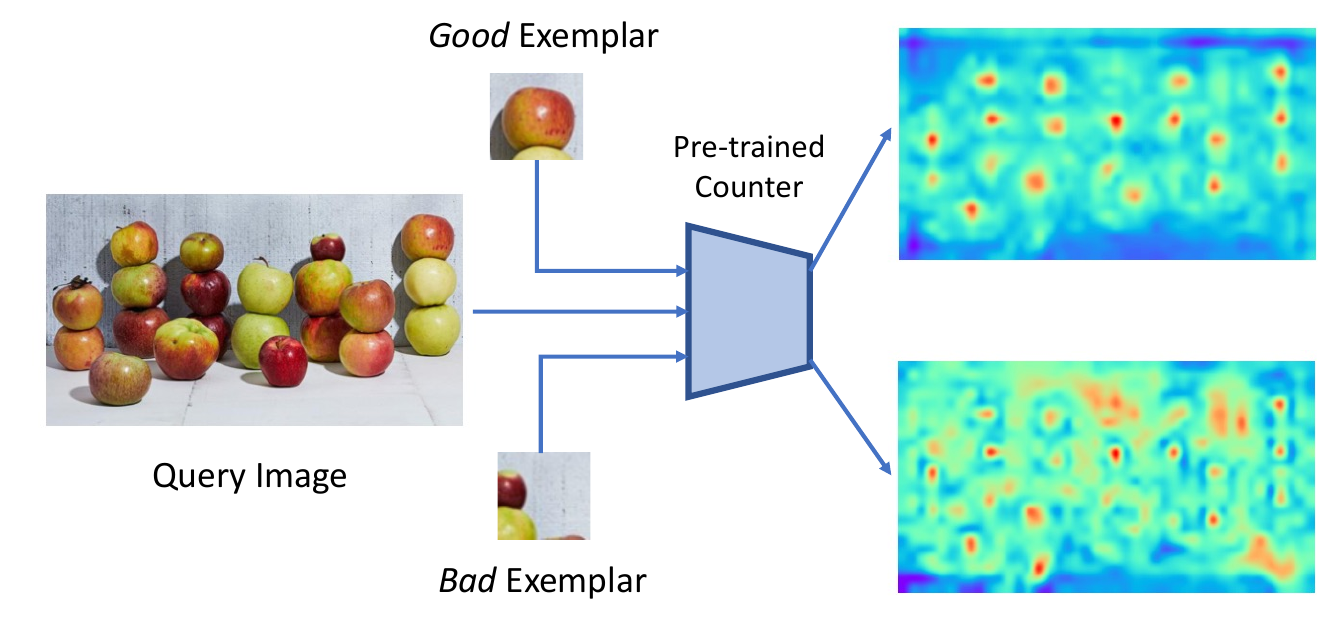} 
\end{center} \vspace{2mm}
\caption{
Feature maps obtained using different exemplars given a pre-trained exemplar-based counting model. The feature maps obtained using good exemplars typically exhibit some repetitive patterns while the patterns from bad exemplars are more irregular.
} \vspace{1mm}
\label{fig:teaser2}
\end{figure}

After obtaining the class-relevant patches, we want to select among them the optimal patches to be used as counting exemplars. Here we observe that the feature maps obtained using \textit{good} 
exemplars and \textit{bad} exemplars often exhibit distinguishable differences. 
An example of the feature maps obtained with different exemplars is shown in Figure \ref{fig:teaser2}. The feature map from a \textit{good} exemplar typically exhibits some repetitive patterns (e.g., the dots on the feature map) that center around the object areas while the patterns from a \textit{bad} exemplar are more irregular and occur randomly across the image. Based on this observation, we train a model to measure the goodness of an input patch based on its corresponding feature maps. Specifically, given an arbitrary patch and a pre-trained exemplar-based object counter, we train this model to predict the counting error of the counter when using the patch as the exemplar. Here the counting error can indicate the goodness of the exemplar.
After this error predictor is trained, we  use it to select those patches with the smallest predicted errors as the final exemplars for counting.

Experiments on the FSC-147 dataset show the effectiveness of our proposed patch selection method. We also provide analyses to show that patches selected by our method can be used in other exemplar-based counting methods to achieve exemplar-free counting. In short, our main contributions are: 
\begin{itemize}
\setlength\itemsep{.3em}
\item We introduce the task of zero-shot object counting that counts the number of instances of a specific class in the input image,  given only the class name and without relying on any human-annotated exemplars.
\item We leverage language-vision models to construct class prototypes via two approaches: VAE-based approach and SD-based approach. We show that in both cases the class prototypes can be used to accurately select patches containing objects of interests for counting.
\item We introduce an error prediction model to further select the optimal patches that yield the smallest counting errors.
\item We verify the effectiveness of our patch selection method on the FSC-147 dataset, through extensive ablation studies and visualization results.

\end{itemize}

\section{Related Work}
\label{sec:rw}
\vspace{1em}
\subsection{Class-specific Object Counting} Class-specific object counting focuses on counting pre-defined categories, such as humans \cite{Lian2019DensityMR,Zhang2019AttentionalNF,Wang2021NWPUCrowdAL,Sindagi2019PushingTF,Idrees2018CompositionLF,Abousamra2021LocalizationIT,Sam2022SSCrowd,Zhang2022CaliFree,Xiong2022DiscreteConstrainedRF,Liu2022LeveragingSF,Wan2021AGL}, animals \cite{Arteta2016CountingIT}, cells \cite{Xie2018MicroscopyCC}, or cars \cite{Mundhenk2016ALC,Hsieh2017DroneBasedOC}. Generally, existing methods can be categorized into two groups: detection-based methods \cite{Chattopadhyay2017CountingEO,Hsieh2017DroneBasedOC,Laradji2018WhereAT} and regression-based methods \cite{Zhang2015CrosssceneCC,Cholakkal2019ObjectCA,Cholakkal2022TowardsPS, Wang2020DMCrowd,zhang2016singleCC,Chan2008PrivacyPC, Liu2019ContextAwareCC}. Detection-based methods apply an object detector on the image and count the number of objects based on the detected boxes. Regression-based methods predict a density map for each input image, and the final result is obtained by summing up the pixel values. Both types of methods require abundant training data to learn a good model. Class-specific counters can perform well on trained categories. However, they can not be used to count objects of arbitrary categories at test time.

\subsection{Class-agnostic Object Counting} 
Class-agnostic object counting aims to count arbitrary categories given only a few exemplars \cite{Lu2018CAC,Ranjan2021LearningTC, Yang2021ClassagnosticFO,Shi2022SimiCounting,Gong2022ClassIntra,Nguyen2022fsoc,Liu2022CounTRTG,fsoc2023you,Arteta2014InteractiveOC}.
GMN \cite{Lu2018CAC} uses a shared embedding module to extract feature maps for both query images and exemplars, which are then concatenated and fed into a matching module to regress the object count.  
FamNet \cite{Ranjan2021LearningTC} adopts a similar way to do correlation matching and further applies test-time adaptation. 
These methods require human-annotated exemplars as inputs. Recently, exemplar-free object counting has been proposed to eliminate the need for user inputs.
Ranjan \textit{et al.} have proposed RepRPN \cite{Ranjan2022Exemplar}, which achieves exemplar-free counting by identifying exemplars from the most frequent objects via a Region Proposal Network (RPN)-based \cite{Ren2015FasterRT} model. However, the class of interest can not be explicitly specified for the RepRPN. In comparison, our proposed method can count instances of a specific class given only the class name.

\subsection{Zero-shot Image Classification} 

Zero-shot classification aims to classify unseen categories for which data is not available during training \cite{Chen2018ZeroShotVR,Jayaraman2014ZeroshotRW,Frome2013DeViSEAD,Rezaei2020ZeroshotLA,RomeraParedes2015AnES,Atzmon2019AdaptiveCS}. Semantic descriptors are mostly leveraged as a bridge to enable the knowledge transfer between seen and unseen classes. Earlier zero-shot learning (ZSL) works  
relate the semantic descriptors with visual features in an embedding space and recognize unseen samples by searching their nearest class-level semantic descriptor in this embedding space \cite{Lampert2009LearningTD,RomeraParedes2015AnES,Xian2016LatentEF,Zhang2017LearningAD}.
Recently, generative models \cite{JingyiICCV21,Xu2022GeneratingRS,m_Le-etal-ECCV18} have been widely employed to synthesize unseen class data to facilitate ZSL \cite{Xian2019FVAEGAND2AF,Xian2019ZeroShotLC,Narayan2020LatentEF}. Xian \textit{et al.} \cite{Xian2019ZeroShotLC} use a conditional Wasserstein Generative Adversarial Network (GAN) \cite{Arjovsky2017WassersteinG} to generate unseen features which can then be used to train a discriminative classifier for ZSL.  
In our method, we also train a generative model conditioned on class-specific semantic embedding. Instead of using this generative model to  hallucinate data, we use it to compute a prototype for each class. This class prototype is then used to select patches that contain objects of interest.

\subsection{Diffusion Models} 

Diffusion models \cite{Ho2020DDPM,Jascha2015Unsupervised,Song2021Score} recently have demonstrated great success in text-to-image generative systems (e.g., DALL-E \cite{Ramesh2022DALLE}, Imagen \cite{Saharia2022Imagen} and Stable Diffusion (SD) \cite{Rombach2022LatentDiff}).
Exploring the potential of pre-trained diffusion models in downstream tasks has gained increasing attention. They have been used in few-shot image classification \cite{Zhang2023StrongFSL}, semantic segmentation \cite{Wu2023Diffu, Karazija2023ZSOpenSeg} and object discovery \cite{Ma2023Diffusionseg}. 
Karazija \etal \cite{Karazija2023ZSOpenSeg} leverage a diffusion-based generative model to produce a set of feature prototypes, which can then be used in a nearest-neighbour lookup scheme to segment images. In our method, we also use a text-conditioned diffusion model (i.e., Stable Diffusion) to construct visual class prototypes. We show that these prototypes can be used to find class-relevant patches for the task of class-agnostic counting.

\begin{figure*}[!ht]
\begin{center}
\includegraphics[width=1.8\columnwidth]{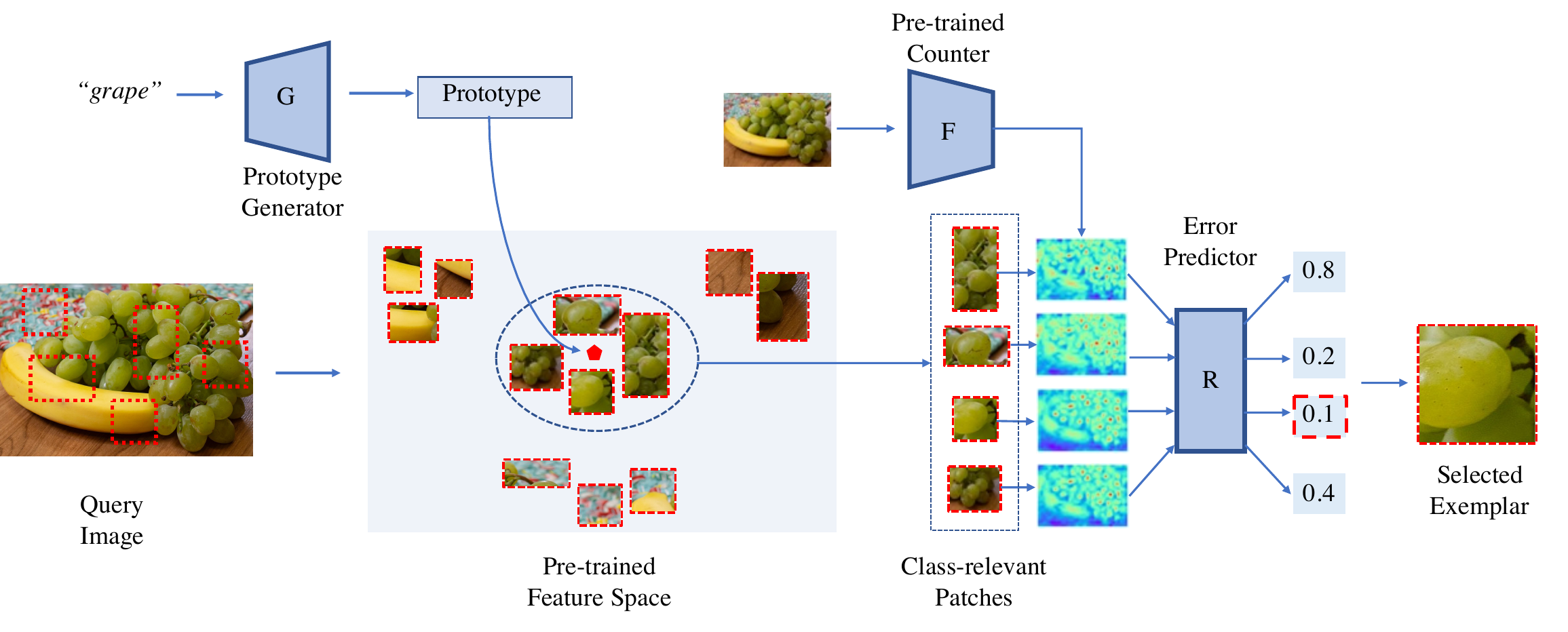} \vspace{-3mm}
\caption{
Overview of the proposed method. We first obtain a class prototype for the given class name ({e.g.} grape) in a pre-trained feature space. Then given an input query image, we generate a set of object proposals with a pre-trained RPN and crop the corresponding image patches. We extract the feature embedding for each patch and select the patches whose embeddings are the nearest neighbors of the class prototype as class-relevant patches. Then for each selected class-relevant patch, we use a pre-trained exemplar-based counting model to obtain the intermediate feature maps. Our proposed error predictor then takes the feature maps as input and predicts the counting error (here we use normalized counting errors). We select the patches with the smallest predicted errors as the final exemplar patches and use them for counting.
} \vspace{-4mm}
\label{fig:overview}
\end{center}

\end{figure*} 

\section{Method}
\label{sec:method}

Figure \ref{fig:overview} summarizes our proposed method. We first construct a class prototype for the given class name in a pre-trained feature space. 
Then given an input query image, we generate a set of object proposals with a pre-trained RPN and crop the corresponding image patches. We extract the feature embedding for each patch and select the patches whose embeddings are the nearest neighbors of the class prototype as class-relevant patches. 
We further use an error predictor to select the patches with the smallest predicted errors as the final exemplars for counting.
We use the selected exemplars in an exemplar-based object counter to infer the object counts. For the rest of the paper, we denote this exemplar-based counter  as the ``base counting model".
We will first describe how we train this base counting model and then present the details of our patch selection method.

\subsection{Training Base Counting Model} 
\label{sec:baseline_training}

We train our base counting model using abundant training images with annotations. Similar to previous works \cite{Ranjan2021LearningTC,Shi2022SimiCounting}, the base counting model uses the input image and the exemplars to obtain a density map for object counting. The model consists of a feature extractor $F$ and a counter $C$. Given a query image $I$ and an exemplar $B$ of an arbitrary class $c$, we input $I$ and $B$ to the feature extractor to obtain the corresponding output, denoted as $F(I)$ and $F(B)$ respectively. $F(I)$ is a feature map of size $d * h_I * w_I $ and $F(B)$ is a feature map of size $d * h_B * w_B $. We further perform global average pooling on $F(B)$ to form a feature vector $b$ of $d$ dimensions.

After feature extraction, we  obtain the similarity map $S$ by correlating the exemplar feature vector $b$ with the image feature map $F(I)$. 
Specifically, if $w_{ij} = F_{ij}(I)$ is the channel feature at spatial position $(i,j)$, $S$ can be computed by:
 \begin{equation}\label{eq:simi}
     S_{ij}(I, B) = w_{ij}^T b.
\end{equation}

In the case where $n$ exemplars are given, we  use Eq. \ref{eq:simi} to calculate $n$ similarity maps, and the final similarity map is the average of these $n$ similarity maps.

We then concatenate the image feature map $F(I)$ with the similarity map $S$, and input them into the counter $C$ to predict a density map $D$.
The final predicted count ${N}$ is obtained by summing over the predicted density map ${D}$:
\begin{equation} \label{eq:final_count}
 {N} = \sum_{i,j}D_{(i,j)}, \vspace{-2mm}   
\end{equation}
where ${D}_{(i,j)}$ denotes the density value for pixel $(i,j)$. 
The supervision signal for training the counting model is the $L_2$ loss between the predicted density map and the ground truth density map:

\begin{equation}\label{eq:counting_loss}
L_{\textnormal{count}} = \|D(I, B) - D^{*}(I)\|_2^2, 
\end{equation}
where $D^{*}$ denotes the ground truth density map.

\subsection{Zero-shot Object Counting}

In this section, we describe how we count objects of any unseen category given only the class name without access to any exemplar. Our strategy is to select a few patches in the image that can be used as exemplars for the base counting model. These patches are selected such that: 1) they contain the objects that we are counting and 2) they benefit the counting model, i.e., lead to small counting errors.
\subsubsection{Selecting Class-relevant Patches}

To select patches that contain the objects of interest, we first generate a class prototype based on the given class name. The class prototype can be considered as a class center representing the patch-level feature distribution of the corresponding class in an
embedding space.
Then we use the generated class prototype to select the class-relevant patches from a set of object patches cropped from the testing image.

Specifically, we introduce two ways of generating prototypes, {i.e.}, generating semantics prototypes using conditional VAE and generating visual prototypes using samples from a latent text-to-image diffusion model, i.e., Stable Diffusion. 

\textbf{VAE-based prototype generation.} To generate class prototypes, we train a conditional VAE model to generate patch-level visual features for an arbitrary class based on the semantic embedding of the class. This strategy is inspired by previous zero-shot learning approaches \cite{Xian2019FVAEGAND2AF,Xian2019ZeroShotLC}. The semantic embedding is obtained from a pre-trained text-vision model \cite{Radford2021LearningTV} given the corresponding class name. 
Specifically, we train a VAE model to reconstruct deep features extracted from a pre-trained ImageNet model. The VAE is composed of an Encoder $E$, which maps a visual feature $x$ to a latent code $z$, and a decoder $G$ which reconstructs $x$ from $z$. Both $E$ and $G$ are conditioned on the semantic embedding $a$ .
 The loss function for training this VAE for an input feature $x$ can be defined as:
 \begin{equation}\label{eq:cvae}
 \begin{aligned}
      L_{V}(x) =  \textnormal{KL} \left( q(z|x,a)||p(z|a) \right)  \\
      - \textnormal{E}_{q(z|x, a)}[\textnormal{log }p(x|z,a)].
\end{aligned}
\end{equation}

The first term is the Kullback-Leibler divergence between the VAE posterior $q(z|x,a)$ and a prior distribution $p(z|a)$. The second term is the decoder's reconstruction error. $q(z|x,a)$ is modeled as $E(x, a)$ and $p(x|z,a)$ is equal to $G(z, a)$. The prior distribution is assumed to be $\mathcal{N}(0,I)$ for all classes.

We can use the trained VAE to generate the semantics prototype for an arbitrary target class for counting. Specifically, given the target class name $y$, we first generate a set of features by inputting the respective semantic vector $a^y$ and a noise vector $z$ to the decoder $G$:
\begin{equation}
\mathbb{G}^y = \{ \hat{x} | \hat{x} = G(z, y), z \sim \mathcal{N}(0, I)\}.
\end{equation}
The class prototype $\textnormal{p}^y$ is computed by taking the mean of all the features generated by VAE:  
 \begin{equation}
 \label{eq:prototype}
 \textnormal{p}^y = \frac{1}{|\mathbb{G}^y|} {\sum}_{\hat{x} \in \mathbb{G}^y}
{\hat{x}} 
\end{equation}

\textbf{SD-based prototype generation.} 
In addition to VAE-based approach for prototype generation, we further leverage the recent advancements in text-to-image models, i.e., Stable Diffusion, to construct class prototypes from SD-generated images. Compared to classic VAE-based models, Stable Diffusion enables more realistic and diverse sample generation, which allows handling the intra-class variation among query objects more effectively.
 
\begin{figure}[!t]
\begin{center}
\includegraphics[width=\linewidth]{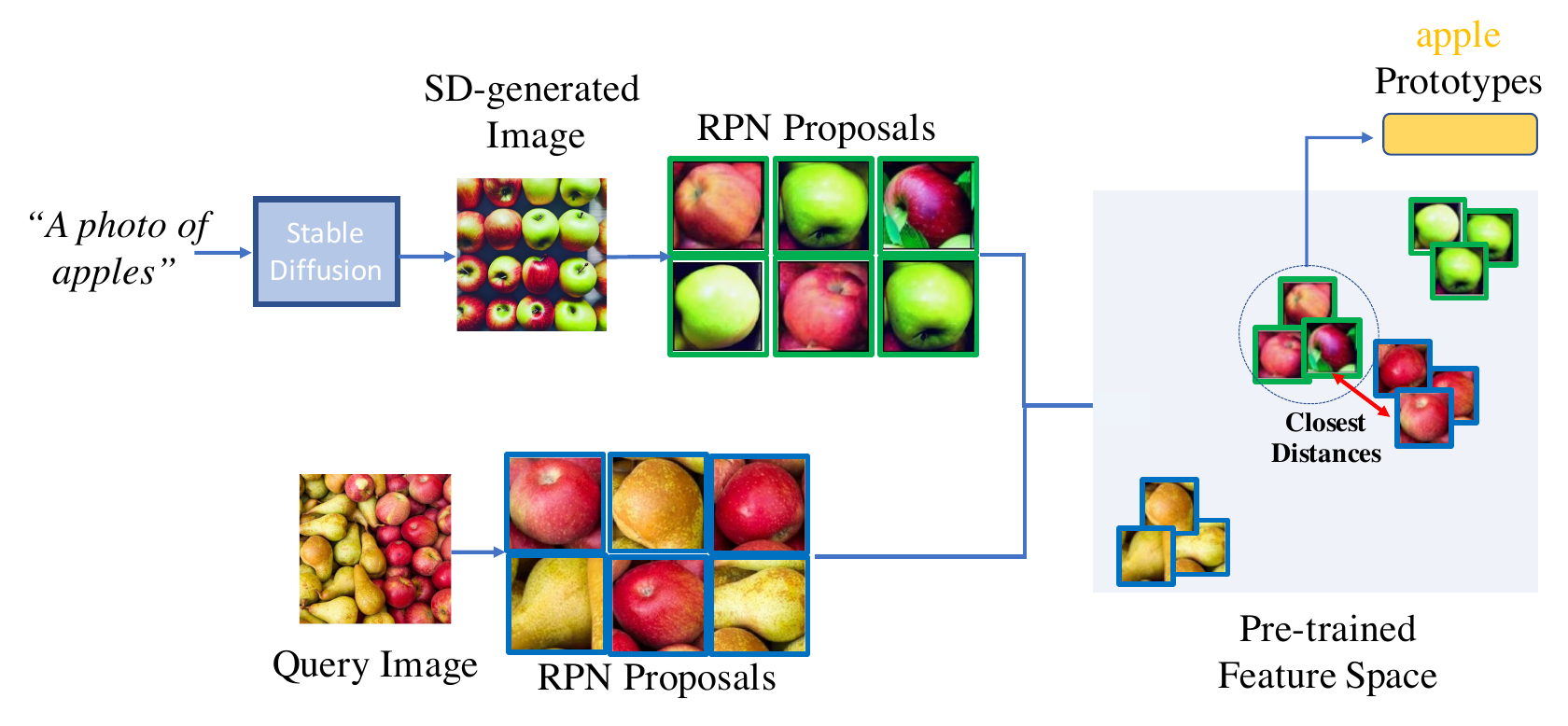} 
\end{center} \vspace{-3mm}
\caption{
Prototype generation using Stable Diffusion.
} \vspace{-2mm}
\label{fig:prototype}
\end{figure}

Specifically, given the target class name for counting, we first use a pre-trained Stable Diffusion model to generate a set of images with the class name as prompt.
We observe that the SD-generated images often contain multiple object instances in various contexts and backgrounds. However, our goal is to obtain a few representative object crops of the target class that can be used to construct reference prototypes. In particular, given a query image, we aim to find a few diffusion-generated object crops that most resemble the target objects in the query image. To do so, we first apply a pre-trained RPN to predict object proposals on both the diffusion-generated images and the query image. Then we compute the pairwise distance between the diffusion-generated object embeddings and the query image's object embeddings. We select the top-$k$ diffusion-generated object embeddings with the nearest mean distance over all query embeddings. We average these $k$ embeddings to construct the visual class prototype. An example is shown in Figure \ref{fig:prototype} where the target objects are red apples. We first obtain a set of object patches containing various crops of apples and select from them a set of red apples to construct the prototype. 

\textbf{Class-relevant patch selection.} 
Using the class prototype, either generated using VAE or Stable Diffusion, we can select the class-relevant patches across the query image. Specifically, we first use a pre-trained RPN to predict object proposals across the query image and extract their corresponding ImageNet features $\{f_1, f_2, ... , f_m\}$. To select the class-relevant patches, we calculate the $L_2$ distance between the class prototype and the patch embedding, namely $d_i = \| f_i - \text{p}^y\|_2$. Then we select the patches whose embeddings are the nearest neighbors of the class prototype as the class-relevant patches.
Since the ImageNet feature space is highly discriminative, {i.e.}, features close to each other typically belong to the same class, the selected patches are likely to contain the objects of the target class.

\subsubsection{Selecting Exemplars for Counting}
Given a set of class-relevant patches and a pre-trained exemplar-based object counter, we aim to select a few exemplars from these patches that are optimal for counting. To do so, we introduce an error prediction network that predicts the counting error of an arbitrary patch when the patch is used as the exemplar. The counting error is calculated from the pre-trained counting model. 
Specifically, to train this error predictor, given a query image $\bar{I}$ and an arbitrary patch $\bar{B}$ cropped from $\bar{I}$, we first use the base counting model to get the image feature map $F(\bar{I})$, similarity map $\bar{S}$, and the final predicted density map $\bar{D}$. The counting error of the base counting model can be written as:  
\begin{equation}
\label{eq:error}
\epsilon = | \sum_{i,j} \bar{D}_{(i,j)} - \bar{N^*}|, \vspace{-2mm}
\end{equation}
where $\bar{N^*}$ denotes the ground truth object count in image $\bar{I}$. 
$\epsilon$ can be used to measure the goodness of $\bar{B}$ as an exemplar for $\bar{I}$, {i.e.}, a small $\epsilon$ indicates that $\bar{B}$ is a suitable exemplar for counting and vice versa.

The error predictor $R$ is trained to regress the counting error produced by the base counting model. The input of $R$ is the channel-wise concatenation of the image feature map $F(\bar{I})$ and the similarity map $\bar{S}$. The training objective is the minimization of the mean squared error between the output of the predictor $R(F(\bar{I}), \bar{S})$ and the actual counting error produced by the base counting model $\epsilon$.

After the error predictor is trained, we can use it to select the optimal patches for counting. The candidates for selection here are the class-relevant patches selected by the class prototype in the previous step. For each candidate patch, we use the trained error predictor to infer the counting error when it is being used as the exemplar. The final selected patches for counting are the patches that yield the top-$s$ smallest counting errors.

\subsubsection{Using the Selected Patches as Exemplars}
Using the error predictor, we predict the error for each candidate patch and select the patches that lead to the smallest counting errors. The selected patches can then be used as exemplars for the base counting model to get the density map and the final count. We also conduct experiments to show that these selected patches can serve as exemplars for other exemplar-based counting models to achieve exemplar-free class-agnostic counting. 
\section{Experiments}
\label{sec:exp}

\subsection{Implementation Details}

\textbf{Network Architecture.} For the \textit{base counting model}, we use ResNet-50 as the backbone of the feature extractor, initialized with the weights of a pre-trained ImageNet model. The backbone outputs feature maps of $1024$ channels. For each query image, the number of channels is reduced to $256$ using an $1 \times 1$ convolution. For each exemplar, the feature maps are first processed with global average pooling and then linearly mapped to obtain a $256$-d feature vector. The counter consists of $5$ convolutional and bilinear upsampling layers to regress a density map of the same size as the query image. 
For the \textit{feature generation model}, both the encoder and the decoder are two-layer fully-connected (FC) networks with 4096 hidden units. LeakyReLU and ReLU are the non-linear activation functions in the hidden and output layers, respectively. The dimensions of the latent space and the semantic embeddings are both set to be $512$.
The \textit{error predictor} composes of $5$ convolutional and bilinear upsampling layers, followed by a linear layer to output the counting error.

\textbf{Dataset.} We use the FSC-147 dataset \cite{Ranjan2021LearningTC} to train the base counting model and the error predictor. FSC-147 is the first large-scale dataset for class-agnostic counting. It includes $6135$ images from $147$ categories varying from animals, kitchen utensils, to vehicles. The categories in the training, validation, and test sets do not overlap. The feature generation model is trained using ImageNet features extracted from MS-COCO objects. 

\textbf{Training Details.} Both the base counting model and the error predictor are trained using the AdamW optimizer with a fixed learning rate of $10^{-5}$. The base counting model is trained for $300$ epochs with a batch size of $8$. We resize the input query image to a fixed height of $384$, and the width is adjusted accordingly to preserve the aspect ratio of the original image.
Exemplars are resized to $128 \times 128$ before being input into the feature extractor. 
To select the class-relevant patches, we use the Region Proposal Network of Faster RCNN pre-trained on MS-COCO dataset to generate $100$ object proposals per image.
The feature generation model is trained using the Adam optimizer and the learning rate is set to be $10^{-4}$. With the VAE-based generator, the semantic embeddings are extracted from CLIP \cite{Radford2021LearningTV}. For visual prototype generation, we use stable-diffusion-v1-4 \cite{Rombach2022LatentDiff} pre-trained on the laion dataset \cite{Schuhmann2022LAION5BAO}. The generated images are in the size of $512$ × $512$. For each generated image, we take the top-$5$ RPN proposals with the highest objectness scores. We combine all the proposals from diffusion-generated images and extract their embeddings to do similarity matching with the object embeddings from the query image. We select the top-$5$ embeddings with nearest distances over query embeddings and compute their average to obtain the class prototype.
The final selected patches are those that yield the top-$3$ smallest counting errors predicted by the error predictor.

 \def\subboxsize{0.45\textwidth}
 \begin{figure*}[ht!]
 \centering
\hspace*{-0.1cm}\includegraphics[width=0.8\linewidth]{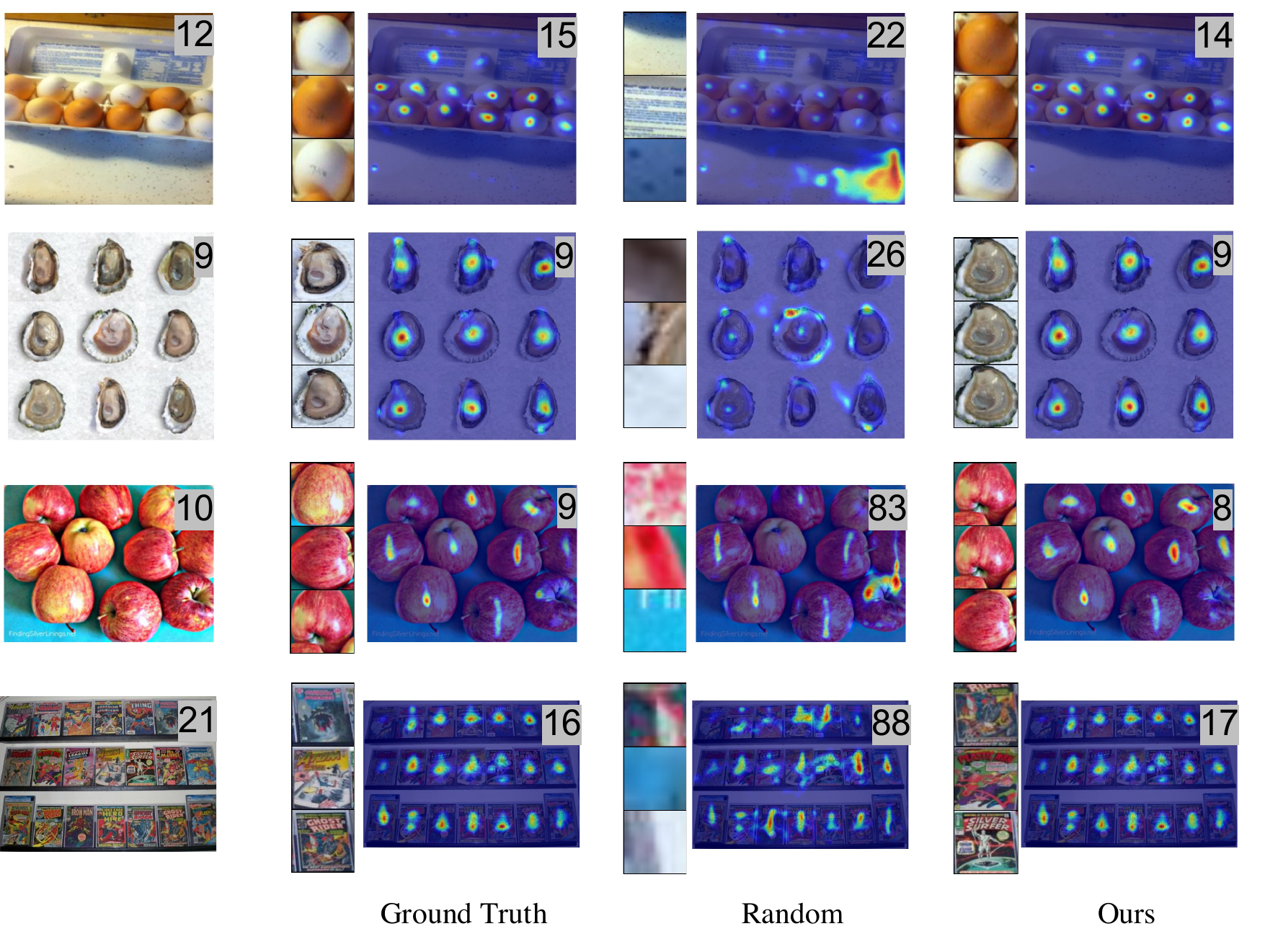}
 \caption{Qualitative results on the FSC-147 dataset. We show the counting exemplars and the corresponding density maps of ground truth boxes, randomly selected patches, and our selected patches respectively. Predicted counting results are shown at the top-right corner. Our  method  accurately identifies suitable patches for counting and the predicted density maps are close to the ground truth density maps.
} \vspace{-2mm}
\label{fig:img_visualization}
\end{figure*}

\subsection{Evaluation Metrics}

We use Mean Average Error (MAE) and Root Mean Squared Error (RMSE) to measure the performance of different object counters. Besides, we follow \cite{Nguyen2022fsoc} to report the Normalized Relative Error (NAE) and Squared Relative Error (SRE). In particular, MAE = $\frac{1}{n} \sum_{i=1}^n |y_i-\hat{y_i}|$; RMSE = $\sqrt{\frac{1}{n} \sum_{i=1}^n (y_i-\hat{y_i})^2}$; NAE = $\frac{1}{n} \sum_{i=1}^n \frac{|y_i-\hat{y_i}|}{y_i}$; SRE = $\sqrt{\frac{1}{n} \sum_{i=1}^n \frac{(y_i-\hat{y_i})^2}{y_i}}$ where $n$ is the number of test images, and $y_i$ and $\hat{y_i}$ are the ground truth and the predicted number of objects for image $i$ respectively. Compared with the absolute errors MAE and RMSE, the relative errors NAE and SRE better reflect the practical usage of visual counting \cite{Nguyen2022fsoc}.

\subsection{Comparing Methods}

We compare our method with the previous works on class-agnostic counting, which can be categorized into exemplar-based counting methods and reference-less counting methods. Exemplar-based methods include FamNet (Few-shot adaptation and matching Network \cite{Ranjan2021LearningTC}), BMNet (Bilinear Matching Network \cite{Shi2022SimiCounting}), CounTR (Counting TRansformer \cite{Liu2022CounTRTG}) and SAFECount (Similarity-Aware Feature Enhancement block for object Counting \cite{fsoc2023you}). These methods require a few human-annotated exemplars as inputs.
Reference-less methods, i.e., RepRPN \cite{Ranjan2022Exemplar} and CounTR \cite{Liu2022CounTRTG}, do not require annotated boxes at test time. Nevertheless, the class of interest can not be specified, which makes them only suitable for counting images with a single dominant object class. Our proposed zero-shot counting, is a new setup which allows the user to specify what to count by simply providing the class name without any exemplar.
We also make exemplar-based methods work in the exemplar-free manner by replacing the human-provided exemplars with the exemplars generated by a pre-trained object detector. Specifically, we use the RPN of Faster RCNN pre-trained on MS-COCO dataset and select the top-$3$ proposals with the highest objectness score as the exemplars.
\begin{table*}[!h] 
  \centering
\resizebox{0.76\textwidth}{!}{%
  \begin{tabular}{l|c|cccc|cccc}
    \toprule
   \multirow{2}{*}{Exemplars} & \multirow{2}{*}{Method} & \multicolumn{4}{c|}{Val Set} & \multicolumn{4}{c}{Test Set} \\
    & & MAE & RMSE & NAE & SRE & MAE & RMSE & NAE & SRE \\
    \midrule
     \multirow{6}{*}{Exemplar-based} 
    & {FamNet+ \cite{Ranjan2021LearningTC}} & {23.75} & {69.07} & 0.52 & 4.25 & {22.08} & {99.54} & 0.44 & 6.45  \\
    & {BMNet \cite{Shi2022SimiCounting}} & {19.06} & {67.95} & 0.26 & 4.39 & {16.71} & {103.31} & 0.26 & 3.32 \\
    & {BMNet+ \cite{Shi2022SimiCounting}} &  {15.74} & {58.53} & {0.27} & {6.57} & 14.62 & 91.83 & 0.25 & 2.74  \\
        & CounTR \cite{Liu2022CounTRTG} & 13.13 & 49.83 & 0.24 & 0.45 & 11.95 & {91.23} & 0.23 & 1.72 \\
        & SAFECount \cite{fsoc2023you} & 14.46 & 51.88 & 0.26 & 0.91 & 13.58 & {91.31} & 0.25 & 1.66 \\
        & Ours  & {18.55} & {61.12} & 0.30 & 3.18 & {20.68} & {109.14} & 0.36 & 7.63 \\
    \midrule
    \multirow{6}{*}{Reference-less} & {RepRPN} \cite{Ranjan2022Exemplar} & {30.40} & {98.73} & - & - & {27.45} & {129.69} & - & - \\
    & CounTR \cite{Liu2022CounTRTG} & 17.40 & 70.33 & 0.34 & 1.64 & 14.12 & {108.01} & 0.29 & 1.93 \\
    & {FamNet+ \cite{Ranjan2021LearningTC}} + RPN & {42.85} & {121.59} & 0.75 & 6.94 & {42.70} & {146.08}  & 0.74 & 7.14 \\
    & {BMNet \cite{Shi2022SimiCounting}} + RPN & {37.26} & {108.54} & {0.42} & {5.43} & 37.22 & 143.13 & 0.41 & 5.31 \\
    & {BMNet+ \cite{Shi2022SimiCounting}} + RPN &  {35.15} & {106.07} & {0.41} & {5.28} & 34.52 & 132.64 & 0.39 & 5.26 \\
    & {SAFECount \cite{fsoc2023you}} + RPN & {34.98} & {107.46} & 0.38 & 5.22 & {33.89} & {139.92} & 0.39 & 5.34  \\
    & Ours + RPN & 32.19 & 99.21 & 0.38 & 4.80 & 29.25 & 130.65 & 0.35 & 4.35 \\
    \midrule
    \multirow{2}{*}{Zero-Shot} 
     & Patch-Sel (VAE)  & 27.47 & 90.85 & 0.37 & 4.52 & 23.14 & 114.40 & 0.34 & 3.95  \\
     & Patch-Sel (SD) & \textbf{26.30} & \textbf{88.80} & \textbf{0.34} & \textbf{4.27} & \textbf{21.53} & \textbf{113.28} & \textbf{0.31} & \textbf{3.61} \\
    \bottomrule
  \end{tabular}} \\ \vspace{2mm}
  \caption{ Quantitative comparisons on the FSC-147 dataset.  ``RPN" denotes using the top-3 RPN proposals with the highest objectness scores as exemplars. ``Patch-Sel (VAE)" and ``Patch-Sel (SD)" denotes using patch selection method with VAE-generated prototypes and SD-generated prototypes respectively.
  }\label{tab:test_val}%
  \vspace{-4mm}
\end{table*}

\begin{figure*}[!h]
\begin{center}
\includegraphics[width=1.5\columnwidth]{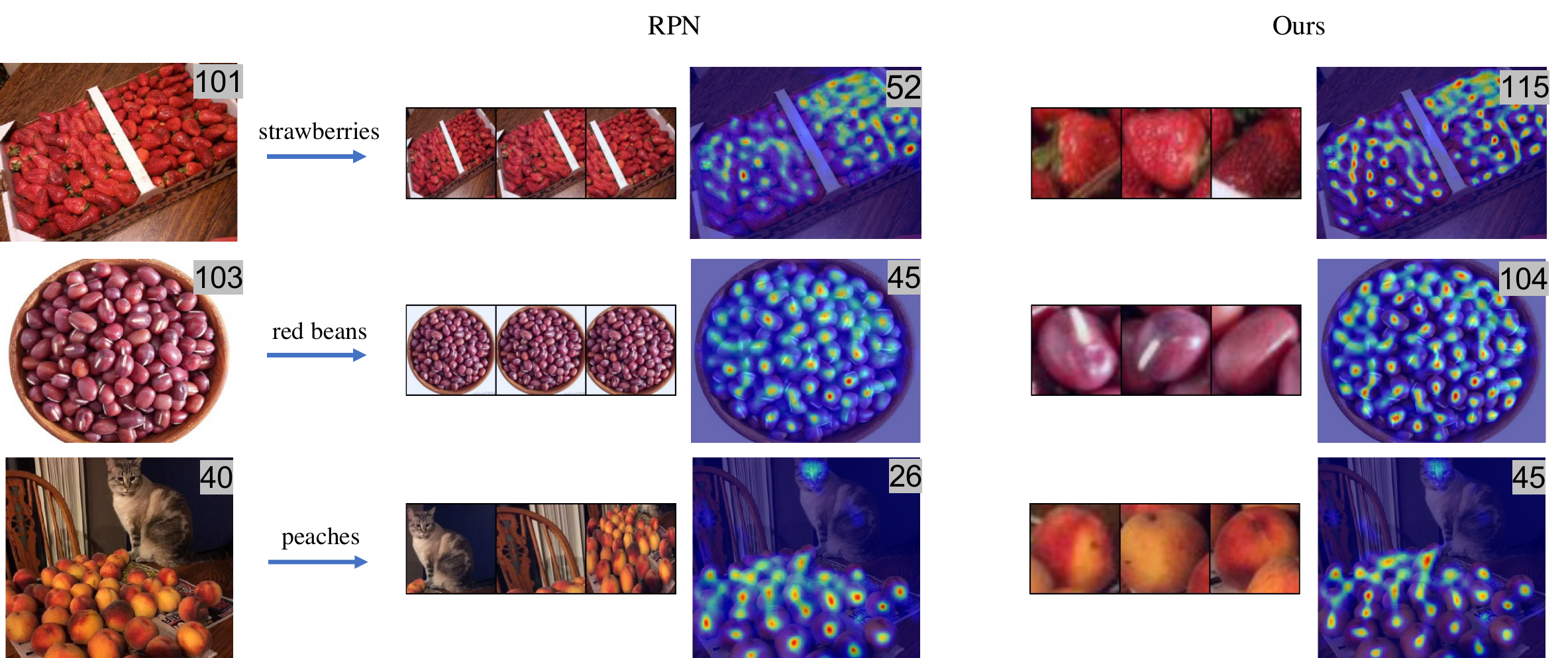} \vspace{0mm}
\caption{
Qualitative comparison with top-$3$ exemplars from RPN. Our proposed method can select patches suitable for counting while RPN-selected patches contain non-relevant objects or multiple object instances.
} 
\label{fig:rpn_comparison}  \vspace{-1.5mm}
\end{center}

\end{figure*} 

\subsection{Results}

\textbf{Quantitative results.} As shown in Table \ref{tab:test_val}, the performance of all exemplar-based counting methods drops significantly when replacing human-annotated exemplars with RPN generated proposals. BMNet+ \cite{Shi2022SimiCounting}, for example, shows an $19.90$ error increase \textit{w.r.t.} the test MAE and a $40.81$ increase \textit{w.r.t.} the test RMSE. 
In comparison, the performance gap is much smaller when using our selected patches as exemplars. Our patch selection method with VAE-generated prototype obtains  
$27.47$ MAE on the validation set and $23.14$ MAE on the test set. By using the SD-generated class prototype, the error rates can be further reduced, achieving $26.30$ MAE on the validation set and $21.53$ MAE on the test set.
Noticeably, compared with the human-annotated exemplars, the NAE and the SRE on the test set are even reduced when using our selected patches.

\textbf{Qualitative analysis.}  In Figure \ref{fig:img_visualization}, we present a few input images, the image patches selected by our method, and the corresponding density maps. Our method effectively identifies the patches that are suitable for object counting. The density maps produced by our selected patches are meaningful and close to the density maps produced by human-annotated patches. The counting model with random image patches as exemplars, in comparison, fails to output meaningful density maps and infers incorrect object counts.

In Figure \ref{fig:rpn_comparison}, we visualize some images from the FSC-147 dataset and the corresponding patches selected by RPN and our method respectively. The RPN-selected patches are the top-$3$ proposals with the highest objectness scores. As can be seen from the figure, the patches selected by RPN may contain objects not relevant to the provided class name or contain multiple object instances. These patches are not suitable to be used as counting exemplars and will lead to inaccurate counting results. This suggests that choosing counting exemplars based on objectness score is not reliable. In comparison, our proposed method can accurately localize image patches according to the given class name. These selected patches can then be used as counting exemplars and yield meaningul density maps and reasonable counting results.

 \def\subboxsize{0.45\textwidth}
 \begin{figure*}[!ht]
 \centering
\includegraphics[width=0.77\linewidth]{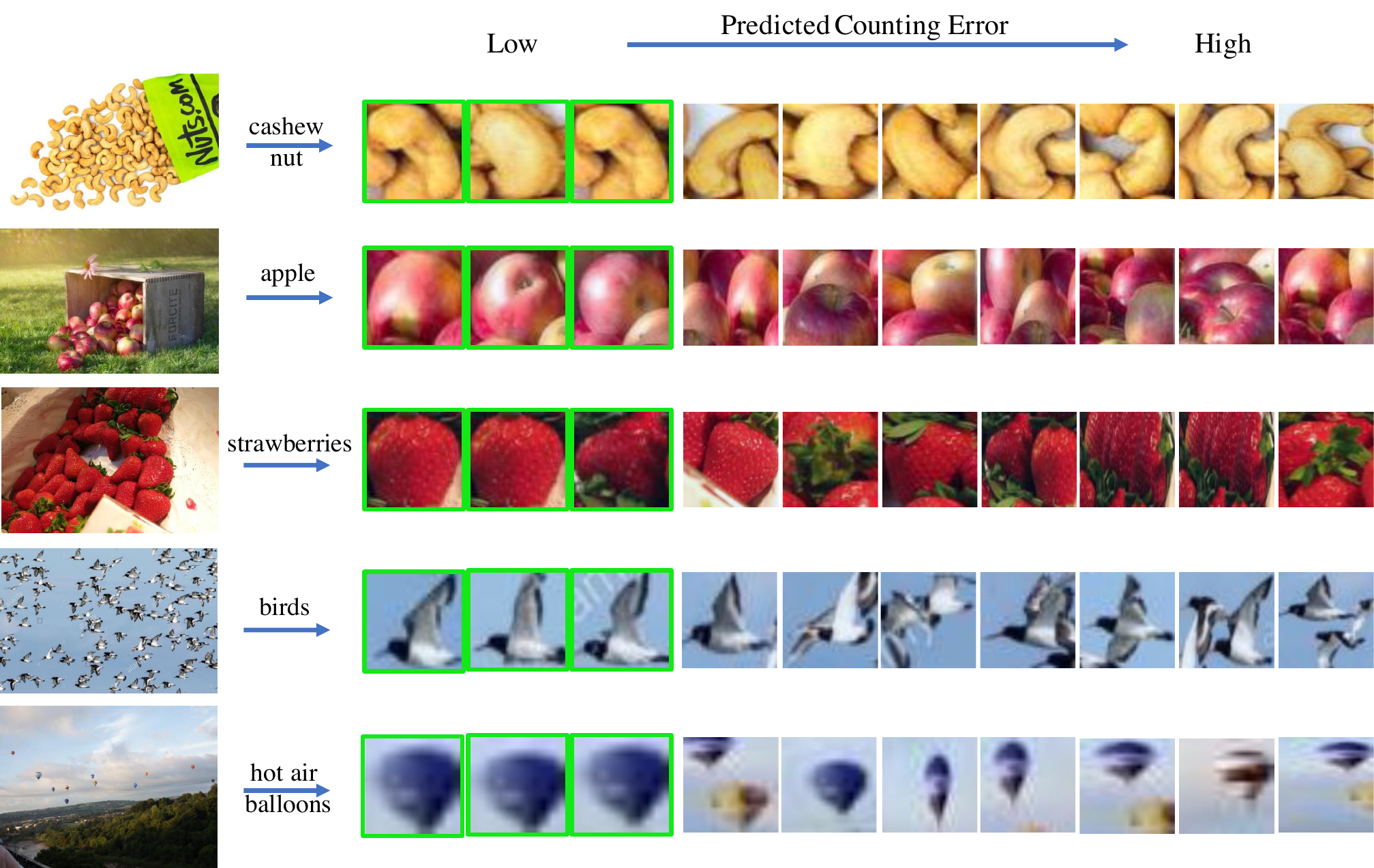} \vspace{0mm}
 \caption{Qualitative ablation analysis. All the $10$ selected class-relevant patches exhibit some class-specific attributes. They are ranked by the predicted counting errors and the final selected patches  with the smallest errors are framed in green.
}
\label{fig:sel_patches}\vspace{2mm}
\end{figure*} 

\section{Analysis}
\subsection{Ablation Studies}
\label{sec:ablation}

Our proposed patch selection method consists of two steps: the selection of class-relevant patches via a generated class prototype and the selection of the optimal patches via an error predictor. We analyze the contribution of each step quantitatively and qualitatively. Quantitative results are in Table \ref{tab:ablation}. We first evaluate the performance of a simple RPN-based baseline, i.e. using  the top-3 RPN proposals with the highest objectness scores as exemplars without any selection step. This baseline method has an error rate of $32.19$ on the validation MAE and $29.25$ on the test MAE. 
As shown in Table \ref{tab:ablation}, using the class prototype generated by VAE to select class-relevant patches reduces the error rate by $2.34$ and $4.87$ on the validation and test set \textit{w.r.t.} MAE, respectively. Using the class prototype generated by Stable Diffusion reduces the MAE by $4.43$ and $7.26$ on the validation and test set respectively. Applying the error predictor can improve the baseline performance by $4.63$ on the validation MAE and $6.46$ on the test MAE. Finally, using the Stable Diffusion prototype and error predictor together further boosts performance, achieving $26.30$ on the validation MAE and $21.53$ on the test MAE. 

We provide further qualitative analysis by visualizing the selected patches. As shown in Figure \ref{fig:sel_patches}, for each input query image, we show $10$ class-relevant patches selected using the class prototype generated with Stable Diffusion, ranked by their predicted counting error (from low to high). All the $10$ selected class-relevant patches exhibit some class specific features. However, not all these patches are suitable to be used as counting exemplars, i.e., some patches only contain parts of the object, and some patches contain some background. By further applying our proposed error predictor, we can identify the most suitable patches with the smallest predicted counting errors. Both the quantitative comparison in Table \ref{tab:test_val} and qualitative results presented in Figure \ref{fig:sel_patches} validate the effectiveness of our patch selection method.

\begin{table}[!h] 
  \centering
\resizebox{0.48\textwidth}{!}{%
  \begin{tabular}{c|c|cccc|cccc}
    \toprule
   \multirow{2}{*}{Prototype}& \multirow{2}{*}{Predictor} & \multicolumn{4}{c|}{Val Set} & \multicolumn{4}{c}{Test Set} \\
    & & MAE & RMSE & NAE & SRE & MAE & RMSE & NAE & SRE \\
    \midrule
    - & - & 32.19 & 99.21 & 0.38 & 4.80 & 29.25 & 130.65 & 0.35 & 4.35 \\
    \Checkmark (VAE) & - & {29.85} & 97.63 & 0.42 & 5.20 & {24.38} & {115.04} & 0.36 & 4.18 \\ 
    \Checkmark (SD) & - & {27.76} & 97.06 & 0.35 & 4.43 & {21.99} & {113.31} & 0.33 & 3.77 \\ 
    - & \Checkmark & {27.56} & \textbf{88.57} & 0.41 & 4.54 & {22.79} & {127.09} & 0.36 & 3.98 \\
    \Checkmark (VAE) & \Checkmark & 27.47 & 90.85 & 0.37 & 4.52 & 23.14 & 114.40 & 0.34 & 3.95  \\
    \Checkmark (SD) & \Checkmark & \textbf{26.30} & {88.80} & \textbf{0.34} & \textbf{4.27} & \textbf{21.53} & \textbf{113.28} & \textbf{0.31} & \textbf{3.61} \\
    \bottomrule
  \end{tabular}} \\ \vspace{2mm}
  \caption{ Ablation study on each component's contribution to the final results. We show the effectiveness of the two steps of our framework: selecting class-relevant patches via a generated class prototype and selecting optimal patches via an error predictor.
  }\label{tab:ablation} \vspace{-3mm}
\end{table}

\begin{table*}[!h] 
  \centering
\resizebox{0.7\textwidth}{!}{%
  \begin{tabular}{l|c|cccc|cccc}
    \toprule
   \multirow{2}{*}{Baseline} & \multirow{2}{*}{Exemplars} & \multicolumn{4}{c|}{Val Set} & \multicolumn{4}{c}{Test Set} \\
    & & MAE & RMSE & NAE & SRE & MAE & RMSE & NAE & SRE \\
        \midrule
    \multirow{3}{*}{FamNet+} & {RPN} & {42.85} & {121.59} & 0.75 & 6.94 & {42.70} & {146.08}  & 0.74 & 7.14 \\
    & Patch-Sel (VAE) & 39.51 & 101.70 & 0.84 & 6.80 & 39.91 & {143.04} & 0.74 & 6.77 \\
    & Patch-Sel (SD) & \textbf{34.60} & \textbf{96.76} & \textbf{0.63} & \textbf{5.71} & \textbf{32.71} & \textbf{139.93}  & \textbf{0.55} & \textbf{5.47} \\
    \midrule
    \multirow{3}{*}{BMNet} & {RPN} & {37.26} & {108.54} & 0.42 & 5.43 & {37.22} & {143.13} & 0.41 & 5.31 \\
    & Patch-Sel (VAE) & 27.71 & 93.98 & 0.36 & 4.55 & 24.45 & {130.42} & 0.33 & 4.09 \\
    & Patch-Sel (SD) & \textbf{26.53} & \textbf{91.55} & \textbf{0.32} & \textbf{4.36} & \textbf{22.27} & \textbf{129.67}  & \textbf{0.28} & \textbf{3.85} \\
    \midrule
    \multirow{3}{*}{BMNet+} & {RPN} & {35.51} & {106.07} & 0.41 & 5.28 & {34.52} & {132.64} & 0.39 & 5.26 \\
    & Patch-Sel (VAE) & 26.89 & 92.37 & 0.35 & 4.56 & 23.14 & {114.40} & 0.34 & 3.95 \\
    & Patch-Sel (SD) & \textbf{25.91} & \textbf{90.62} & \textbf{0.33} & \textbf{4.40} & \textbf{19.45} & \textbf{109.82}  & \textbf{0.27} & \textbf{3.53} \\
    \midrule
    \multirow{3}{*}{SAFECount} & {RPN} & {34.98} & {107.46} & 0.38 & 5.22 & {33.89} & {139.92} & 0.39 & 5.34 \\
    & Patch-Sel (VAE) & 28.34 & 94.22 & 0.41 & 4.65 & 23.60 & \textbf{110.95} & 0.40 & \textbf{4.26} \\
    & Patch-Sel (SD) & \textbf{26.85} & \textbf{91.09} & \textbf{0.33} & \textbf{4.30} & \textbf{21.44} & {115.30}  & \textbf{0.30} & 5.74 \\
    \bottomrule
  \end{tabular}} \\  \vspace{4pt}
  \caption{ Using our selected patches as exemplars for other exemplar-based class-agnostic counting methods (FamNet+, BMNet, BMNet+ and SAFECount) on FSC-147 dataset. Our patch selection method generalizes well to other exemplar-based counting methods.
  }\label{tab:baselines}%
  \vspace{-2mm}
\end{table*}

\subsection{Generalization to Exemplar-based Methods}
\label{sec:dataset}

Our proposed method can be considered as a general patch selection method that is applicable to other visual counters to achieve zero-shot object counting. To verify that, we use our selected patches as the exemplars for four other different exemplar-based methods: FamNet+ \cite{Ranjan2021LearningTC}, BMNet \cite{Shi2022SimiCounting}, BMNet+ \cite{Shi2022SimiCounting} and SAFECount \cite{fsoc2023you}. Table \ref{tab:baselines} shows the results on the FSC-147 dataset. The baseline uses top-3 RPN proposals with the highest objectness scores as exemplars for the pre-trained exemplar-based counter. Our patch selection method with VAE-generated class prototype reduces the error rates by a large margin for all exemplar-based methods. For example, the MAE for BMNet+ \cite{Shi2022SimiCounting} reduces from $35.51$ to $26.89$ on the validation set and from 
$34.52$ to $23.14$ on the test set. By using our patch selection method with the SD-generated class prototype, the error rates are further reduced for most cases, e.g., we observe for FamNet+ \cite{Ranjan2021LearningTC}, there is an error reduction of $12.4\%$ \textit{w.r.t.} the validation MAE and $18.0\%$ \textit{w.r.t.} the test MAE. The consistent performance improvements validate that our patch selection method generalizes well to other exemplar-based counting methods.

\subsection{Multi-class Object Counting}
\begin{figure}[htp]
\subfloat[Broccoli and carrots]{%
  \includegraphics[clip,width=0.96\columnwidth]{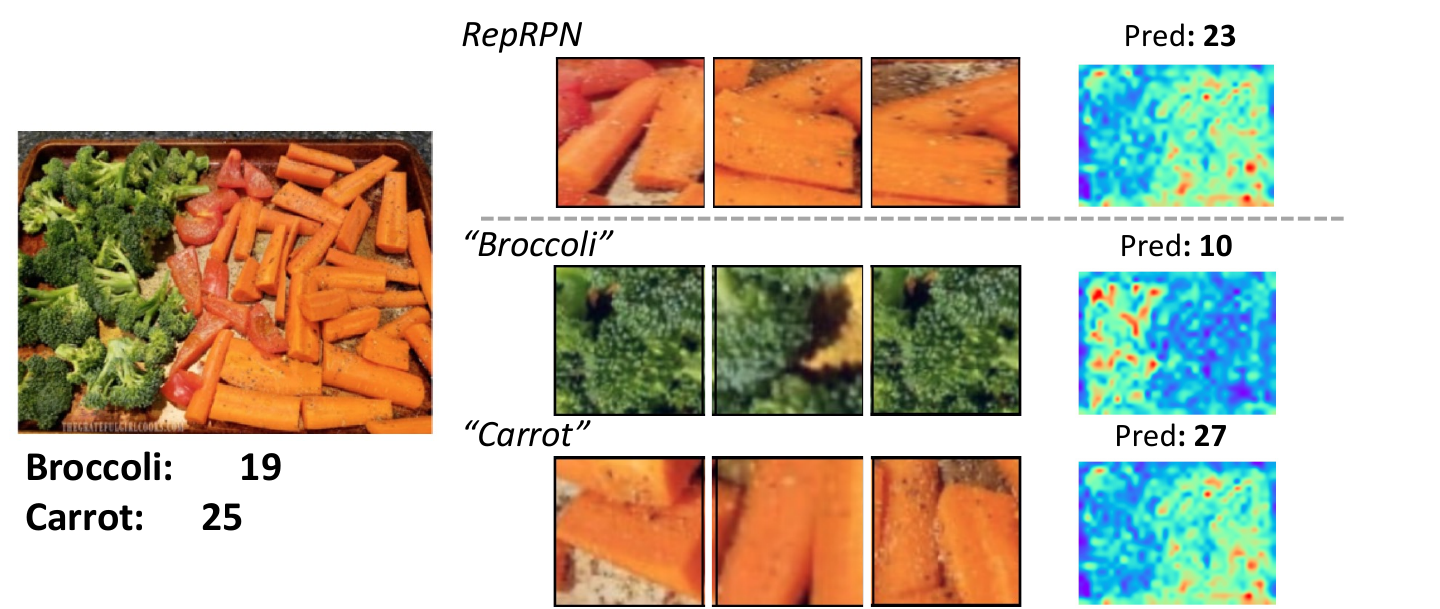}%
}

\subfloat[Bananas and strawberries]{%
  \includegraphics[clip,width=0.96\columnwidth]{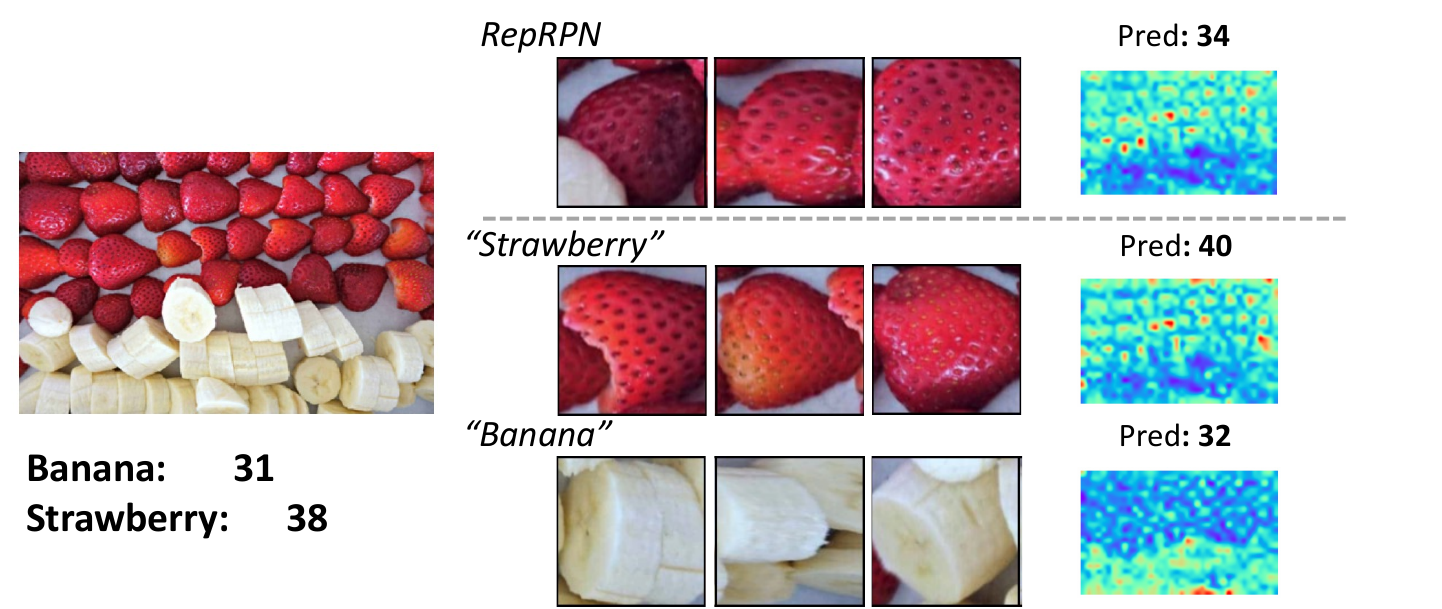}%
}

\subfloat[Green Beans and tomatoes]{%
  \includegraphics[clip,width=0.96\columnwidth]{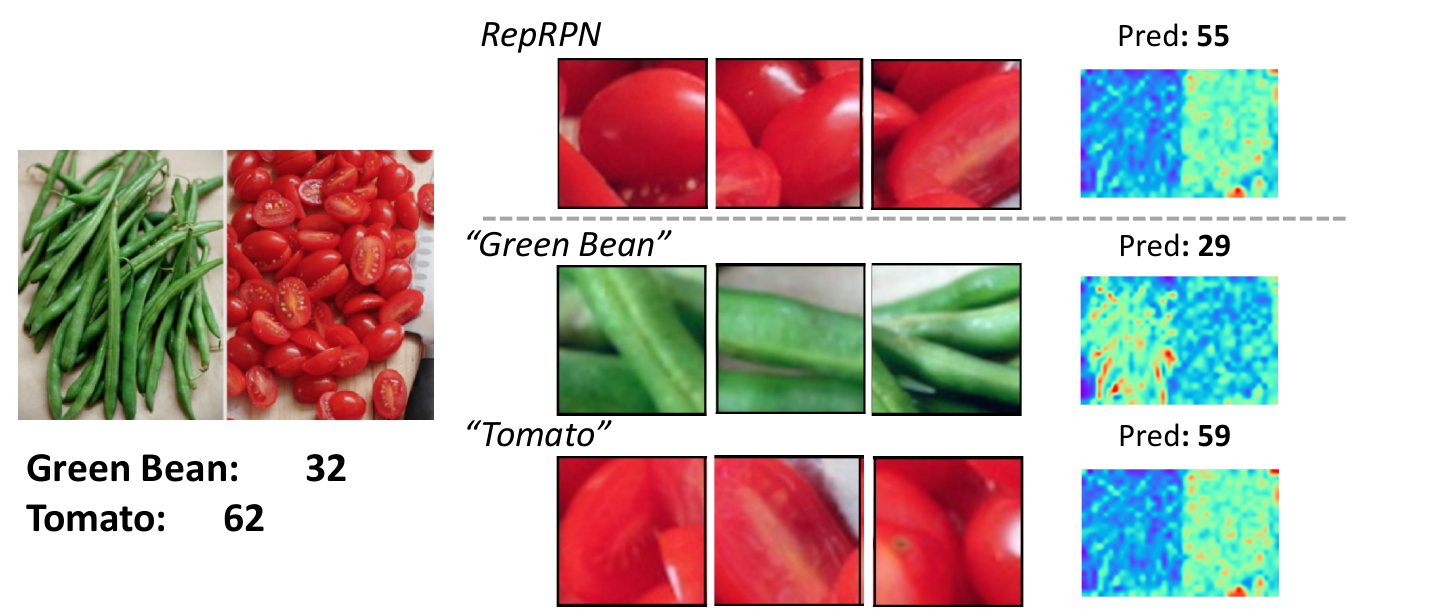}%
}
\caption{Visualization results of our method on some multi-class examples. Our method selects patches according to the given class name and the corresponding heatmap highlights the relevant areas.}\label{fig:multi} \vspace{-2mm}
\end{figure}

Our method can count instances of a specific class given the class name, which is particularly useful when there are multiple classes in the same image. In this section, we show some visualization results in this multi-class scenario. As shown in Figure \ref{fig:multi}, our method selects patches according to the given class name and counts instances from that specific class in the input image. Correspondingly, the heatmap highlights the image regions that are most relevant to the specified class. Here the heatmaps are obtained by correlating the exemplar feature vector with the image feature map in a pre-trained ImageNet feature space. Note that we mask out the image region where the activation value in the heatmap is below a threshold when counting the objects of interests.
We also show the patches selected using another exemplar-free counting method, RepRPN \cite{Ranjan2022Exemplar}. The class of RepRPN selected patches can not be explicitly specified. It simply selects patches from the class with the highest number of instances in the image according to the repetition score.

\subsection{Qualitative Comparison between SD-generated Prototypes and VAE-generated Prototypes}
\begin{figure*}[!ht]
\begin{center}
\includegraphics[width=1.5\columnwidth]{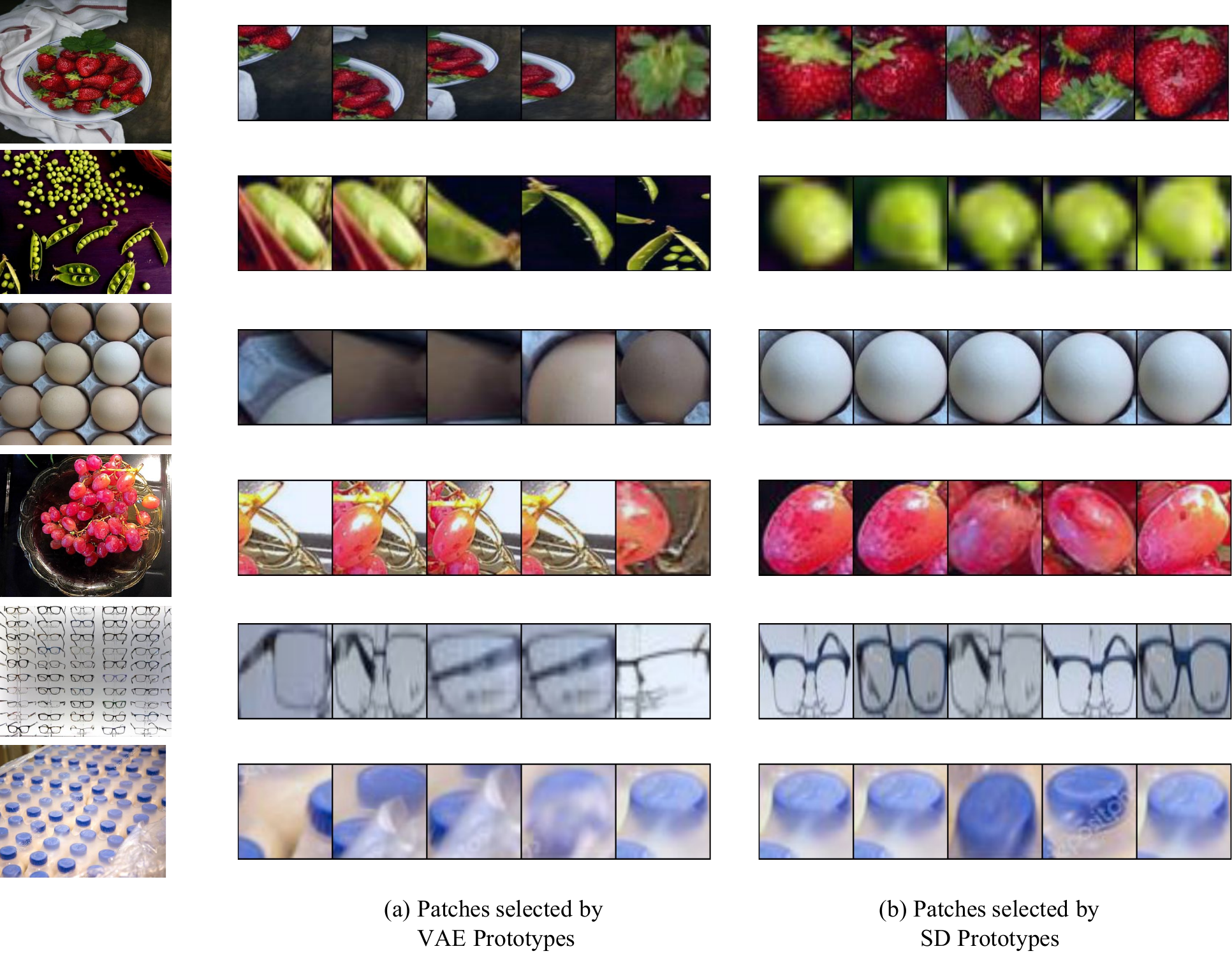} \vspace{0mm}
\caption{
Visualization of the patches selected by VAE-generated prototypes and SD-generated prototypes. Patches selected by SD-generated prototypes are of higher quality. 
} \vspace{-2mm}
\label{fig:vae_sd_comparison}
\end{center}

\end{figure*}

In this section, we provide qualitative comparison between patches selected via SD-generated prototypes and VAE-generated prototypes.
As shown in Figure \ref{fig:vae_sd_comparison}, we present a few input images and the corresponding patches selected by VAE-generated prototypes and SD-generated prototypes. Although the patches selected by VAE-generated prototypes generally contain the objects of interest, they miss parts of the objects in some cases (e.g., the second patch of \textit{grape}), or contain multiple object instances within one patch (e.g., the second patch of \textit{strawberry}). 
In comparison, the patches selected by SD-generated prototypes are generally better exemplars for counting, i.e., one patch mostly contains a single complete object instance. 

\begin{figure*}[htp]
\begin{center}
\subfloat[Grapes]{%
  \includegraphics[clip,width=1.45\columnwidth]{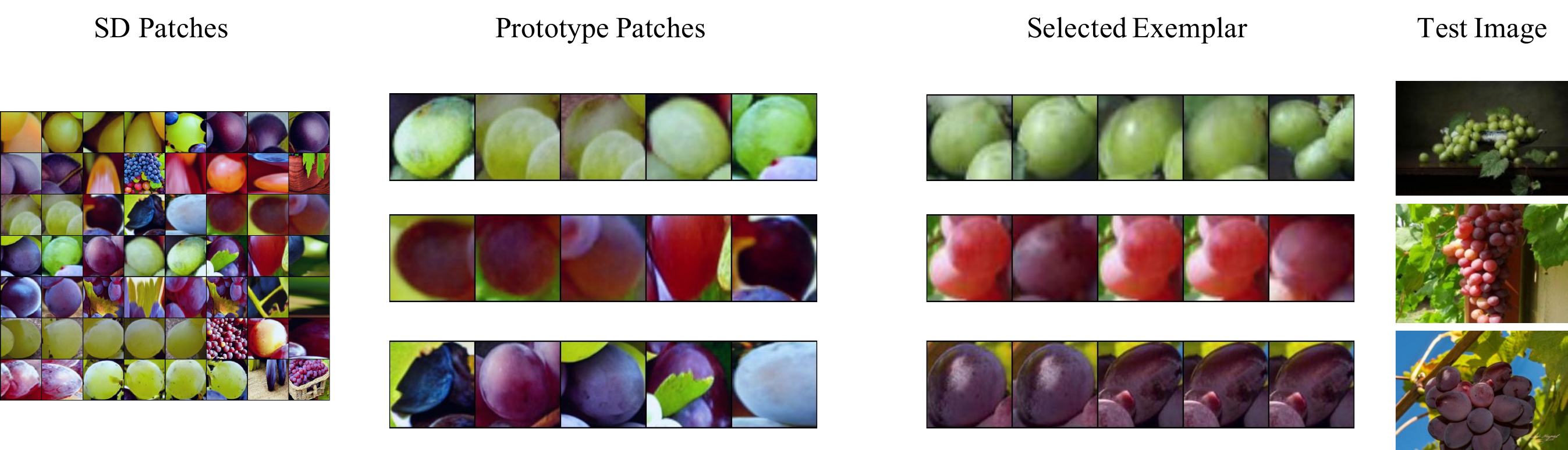}%
}

\subfloat[Eggs]{%
  \includegraphics[clip,width=1.45\columnwidth]{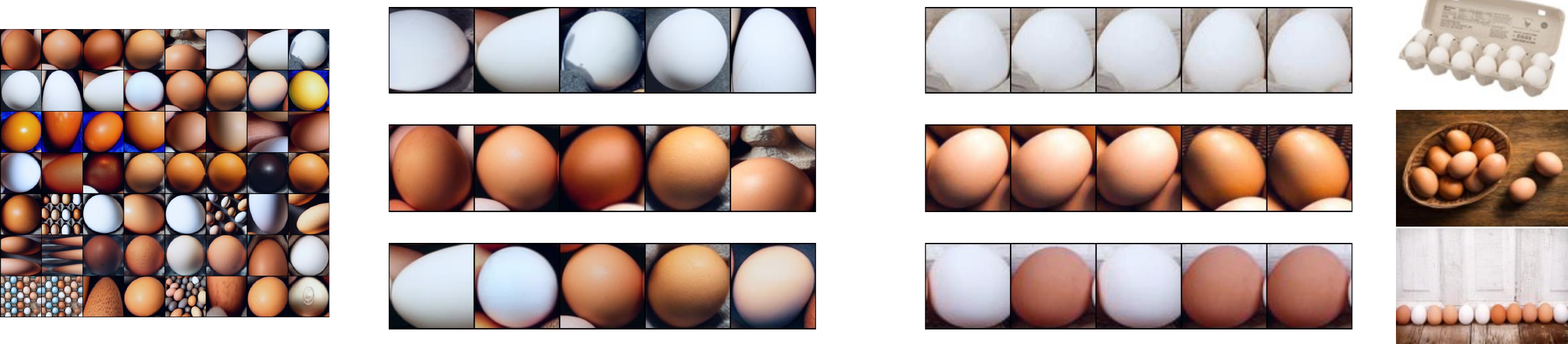}%
}

\subfloat[Apples]{%
  \includegraphics[clip,width=1.45\columnwidth]{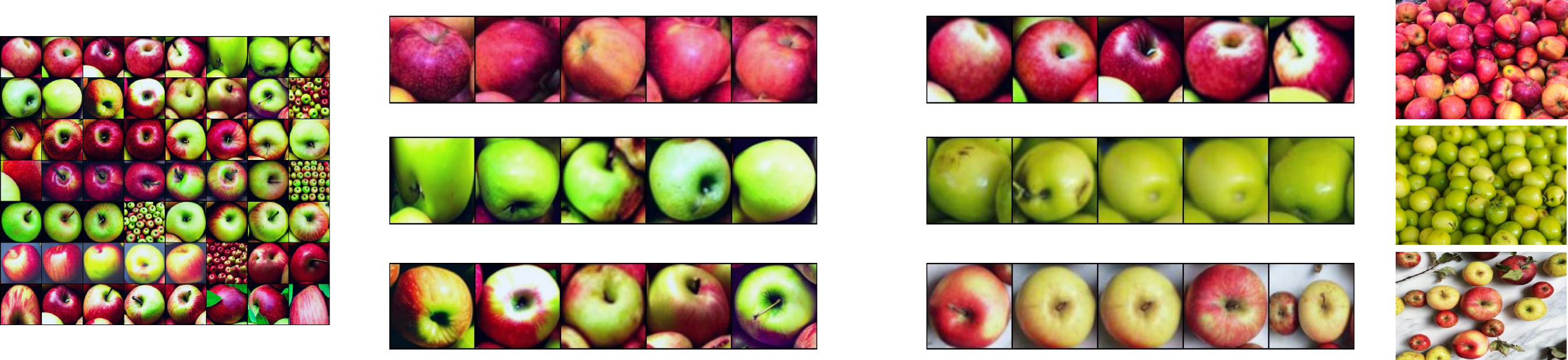}%
}
\caption{Visualization of the diffusion-generated patches for prototype generation and the corresponding selected patches from query images. Our method constructs prototypes according to the query images and selects the patches that most resemble the target objects.}
\label{fig: sd_prototype}
\end{center}
\end{figure*}

\subsection{Analysis on SD-generated Prototypes}

\textbf{Qualitative Visualization.} To generate visual class prototypes, we first use a pre-trained Stable Diffusion to generate a set of object patches for the class of interest. 
Then given a query image, we select the generated patches that most resemble the target objects in the query image. We compute the average feature embeddings of the selected generated patches to construct the prototype.
In Figure \ref{fig: sd_prototype}, we demonstrate this process for three different categories, i.e., grapes, eggs, and apples. For each category, we show how we select different patches to construct the prototypes for the given query image. As can be seen in the first column, the set of RPN proposals extracted from SD-generated images exhibit rich variations. Among these proposals, our method selects those that are most relevant to the testing images for constructing prototypes.
For example, in the last row of Figure \ref{fig: sd_prototype} (c), only apples with mixed colors are selected to match the objects' colors in the testing image.
Compared with VAE-generated prototypes which remain the same for all query images, SD-generated prototypes can better handle the variations of query images and select more accurate counting exemplars. 

\textbf{Number of Patches for Prototype Generation.}
In our main experiments, we select the top-$5$ SD-generated patches with the nearest mean distance over query patches, and compute their average features to construct prototypes. In this section, we conduct an ablation study on how the number of patches selected for prototype generation affects the counting performance. Specifically, we select top-$5$, top-$25$, top-$50$ and all patches to construct class prototypes and use them to select exemplars. Results are summarized in Table \ref{tab:patches_num}. We observe that the performance drops on both the validation set and test set as the number of selected patches increases. The counting errors are highest when using all generated patches to construct prototypes. In this case, the same class prototype is applied for all images of this class, which is not optimal for counting objects with large intra-class diversity. Our method, in comparison, selects the most similar patches based on the query image, which leads to more accurate prototypes.

\begin{table}[!h] 
  \centering
\resizebox{0.3\textwidth}{!}{%
  \begin{tabular}{c|cc|cc}
    \toprule
   {Prototype} & \multicolumn{2}{c|}{Val Set} & \multicolumn{2}{c}{Test Set} \\
    {Patches} &  MAE & RMSE  & MAE & RMSE  \\
    \midrule
    top-5  & \textbf{27.76} & \textbf{97.06} & \textbf{21.99} & \textbf{113.31}  \\
    top-25  & 28.03 & 97.87 & 22.12 & 113.71 \\ 
    top-50  & 28.33 & 98.50 &  22.34 & 113.88 \\ 
    all  & 30.13 & 101.70 & 23.37 & 116.04 \\ 
    \bottomrule
  \end{tabular}} \\ \vspace{2mm}
  \caption{ Ablation study on the number of diffusion-generated patches for prototype generation.
  }\label{tab:patches_num} \vspace{-8mm}
\end{table}

\section{Conclusion} 
We propose a new task, zero-shot object counting, to count instances of a specific class given only the class name without access to any exemplars. To address this, we developed a two-step approach that accurately localizes the optimal patches to be used as counting exemplars. We leverage language-vision models to construct class prototypes via two approaches: VAE-based apporoach and diffusion-based approach. Through these two approaches, we present a comprehensive study of prototype construction at both category and image levels. 
In the context of our specific task, the diffusion-based image prototype has notably outperformed the category-level prototype constructed via VAE, thanks to its ability to customize prototypes to match object appearances in each image. More generally, our approach of employing language-vision models for zero-shot recognition is applicable in various tasks. In scenarios where 
data of large-scale generative models do not exist such as medical imaging, or remote sensing, VAE can be a useful choice. 

We show that the prototypes can be used to select the patches containing objects of interests. 
Furthermore, we introduce an error prediction model to select those patches with the smallest predicted errors as the final exemplars for counting.
Extensive results demonstrate the effectiveness of our method. We also conduct experiments to show that our selected patches can be incorporated into other exemplar-based counting methods to achieve exemplar-free counting.

\ifCLASSOPTIONcompsoc
 
\else
\fi

\ifCLASSOPTIONcaptionsoff
  \newpage
\fi



%

\bibliography{shortstrings,egbib}
\bibliographystyle{IEEEtran}
\vspace{-12mm}
\begin{IEEEbiography}[{\includegraphics[width=1in,height=1.25in,clip,keepaspectratio]{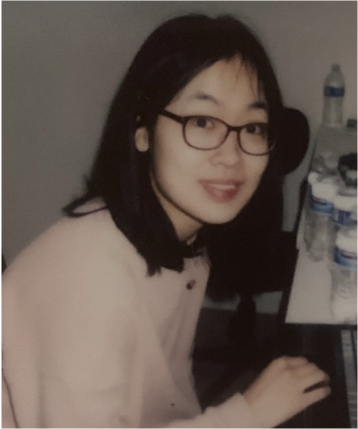}}]{Jingyi Xu}
received her B.S. degree in computer science from Nankai University in 2019. She is currently working toward the PhD degree in computer science with the Stony Brook University. Her research interests include computer vision and deep learning, with a focus on representation learning and few-shot learning.
\end{IEEEbiography}
\vspace{-12mm}
\begin{IEEEbiography}[{\includegraphics[width=1in,height=1.25in,clip,keepaspectratio]{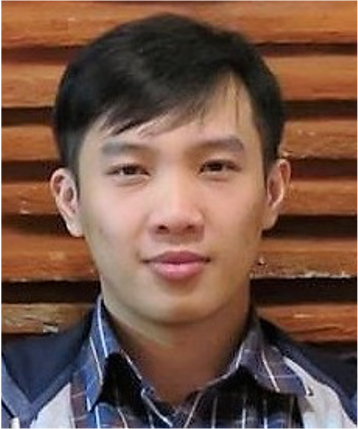}}]{Hieu Le}
received his B.S. degree in computer science from the Ho Chi Minh City University of Science in 2012, and his Ph.D. degree in computer science from Stony Brook University in 2020. He is currently a Postdoc at CVLab, EPFL. His research interest includes illumination modeling, generative models, and 3D modelling. 
\end{IEEEbiography}
\vspace{-12mm}
\begin{IEEEbiography}[{\includegraphics[width=1in,height=1.25in,clip,keepaspectratio]{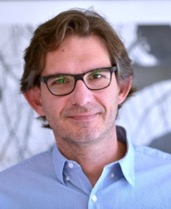}}]{Dimitris Samaras}
received the diploma degree in computer science and engineering from the University of Patras, in 1992, the MSc degree from Northeastern University, in 1994, and the PhD degree from the University of Pennsylvania, in 2001. He is an associate professor of computer science with Stony Brook University. He specializes in illumination modeling and estimation for recognition and graphics,
and biomedical image analysis. He is Program Chair of CVPR 2022.
\end{IEEEbiography}

\end{document}


\onecolumn

\newcommand{\titlename}{ 
Zero-Shot Object Counting \\ with Language-Vision Models \\
Supplementary Material}
\title{\titlename}

\author{Jingyi Xu, Hieu Le and~Dimitris Samaras 
}


\IEEEtitleabstractindextext{

}

\maketitle
\IEEEdisplaynontitleabstractindextext
\IEEEpeerreviewmaketitle

In this document, we provide additional experiments and analyses. In particular:
\begin{itemize}
   \setlength\itemsep{1em}
    \item Section \ref{sec:candidate_patches} provides the performance of our method when using different sets of proposals for patch selection.
    \item Section \ref{sec:add_qual} provides additional qualitative analysis on the class prototypes generated by Stable Diffusion (SD).
     \item Section \ref{sec:comp_clip} provides the comparison between our method and directly using CLIP features for patch selection.
      \item Section \ref{sec:multi_class} provides additional visualizations of our selected patches in multi-class cases.
    \item Section \ref{sec:number_neighbor} provides ablation study on the number of nearest neighbors when selecting class-relevant patches.
    \item Section \ref{sec:failure} provides visualizations on the failure cases when using Stable Diffusion to generate class prototypes.
    
     \item Section \ref{sec:qual_err} provides qualitative analysis of the proposed error predictor. 
\end{itemize}



 \section{Different Methods to Acquire Proposals for Selection}
\label{sec:candidate_patches} 
In our main experiments, the proposals for selection are generated by a pre-trained RPN \cite{Ren2015FasterRT}. In this section, we conduct an ablation study on how to obtain the proposal patches for selection. Instead of using proposals generated via a pre-trained RPN, we use the proposals from selective search \cite{Uijlings2013SS} and a pre-trained Background Aware RPN (BA-RPN) proposed in \cite{Zheng2021zsi}. Results are summarized in Table \ref{tab:sampling}. 
As can be seen from the table, our proposed patch selection method can bring consistent performance improvements for all the three set of proposal patches. with SD-generated prototypes, we observe the lowest MAE on the validation set (i.e., 26.05) using BA-RPN proposals and the lowest MAE on the test set (i.e., 21.53) using RPN
proposals.
\begin{table*}[!ht] 
  \centering
\resizebox{0.75\textwidth}{!}{%
  \begin{tabular}{c|c|cccc|cccc}
    \toprule
   \multirow{2}{*}{\parbox{1.7cm}{Proposals for selection}} & \multirow{2}{*}{Patch Selection} & \multicolumn{4}{c|}{Val Set} & \multicolumn{4}{c}{Test Set} \\
   & & MAE & RMSE & NAE & SRE & MAE & RMSE & NAE & SRE \\
    \midrule
    \multirow{3}{*}{\hspace{9mm}\parbox{1.9cm}{Selective Search \cite{Uijlings2013SS}}} &  \xmark & 33.79 & 102.87 & 0.57 & 6.44 & 29.10 & 133.49 & 0.45 & 4.95  \\ 
       & \checkmark (VAE) & 28.45 & 93.55 & {0.39} & 4.64 & 23.04 & 117.63 & 0.35 & 3.89 \\
       & \checkmark (SD) & 27.70 & 94.77 & {0.36} & 4.46 & 22.68 & 117.96 & 0.32 & 3.92 \\
       \midrule
     \multirow{3}{*}{BA-RPN \cite{Zheng2021zsi}} & \xmark & 32.02 & 97.44 & 0.37 & 4.79 & 26.83 & 130.53 & 0.32 & 4.16 \\
      & \checkmark (VAE) & 26.48 & \textbf{87.67} & 0.34 & {4.22} & {22.22} & {115.04} & {0.32} & {3.70} \\
      & \checkmark (SD) & \textbf{26.05} & 88.39 & \textbf{0.33} & \textbf{4.21} & 21.77 & 116.06 & 0.31 & 3.62 \\
     \midrule
     \multirow{3}{*}{RPN} & \xmark & 32.19 & 99.21 & 0.38 & 4.80 & 29.25 & 130.65 & 0.35 & 4.35 \\
     & \checkmark (VAE) & 27.47 & 90.85 & 0.37 & 4.52 & 23.14 & 114.40 & 0.34 & 3.95 \\
      & \checkmark (SD) &  {26.30} & {88.80} & {0.34} & {4.27} & \textbf{21.53} & \textbf{113.28} & \textbf{0.31} & \textbf{3.61} \\
      
    \bottomrule
  \end{tabular}} \\ \vspace{6pt}
  \caption{ Performance on FSC-147 dataset when using different sets of proposal patches for selection. Our proposed method brings consistent improvement in the performance.
  }\label{tab:sampling}%
\end{table*}

\section{Qualitative Analysis on SD-generated Prototypes}
\label{sec:add_qual} 
\begin{figure*}[htp]
\begin{center}
\subfloat[Flower pots]{%
  \includegraphics[clip,width=0.8\columnwidth]{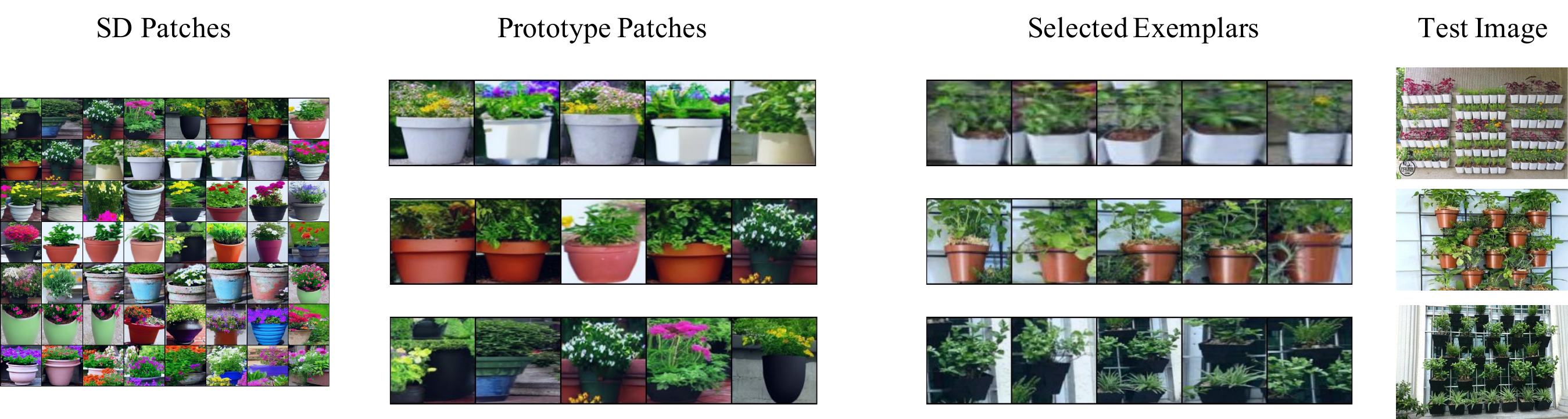}%
}

\subfloat[Pills]{%
  \includegraphics[clip,width=0.8\columnwidth]{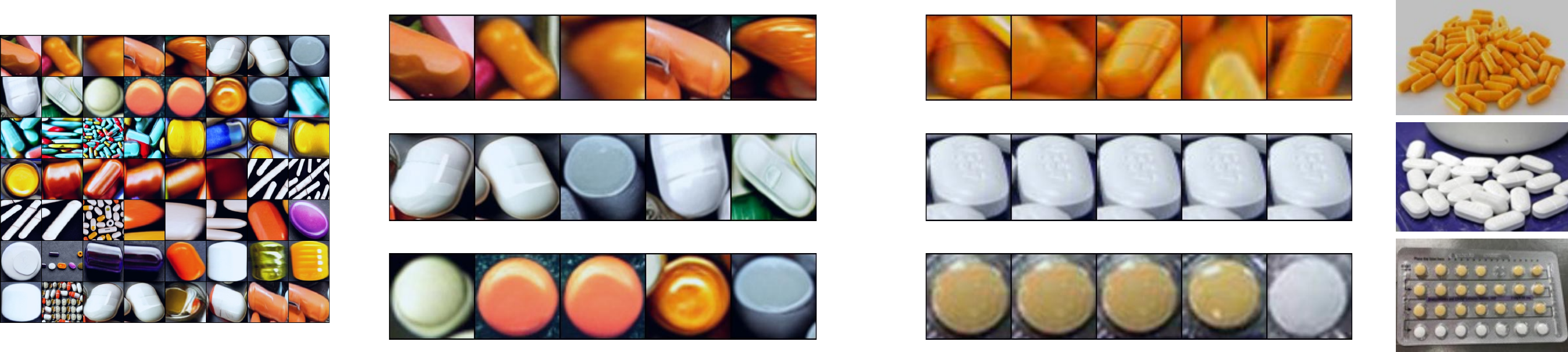}%
}

\subfloat[Marbles]{%
  \includegraphics[clip,width=0.8\columnwidth]{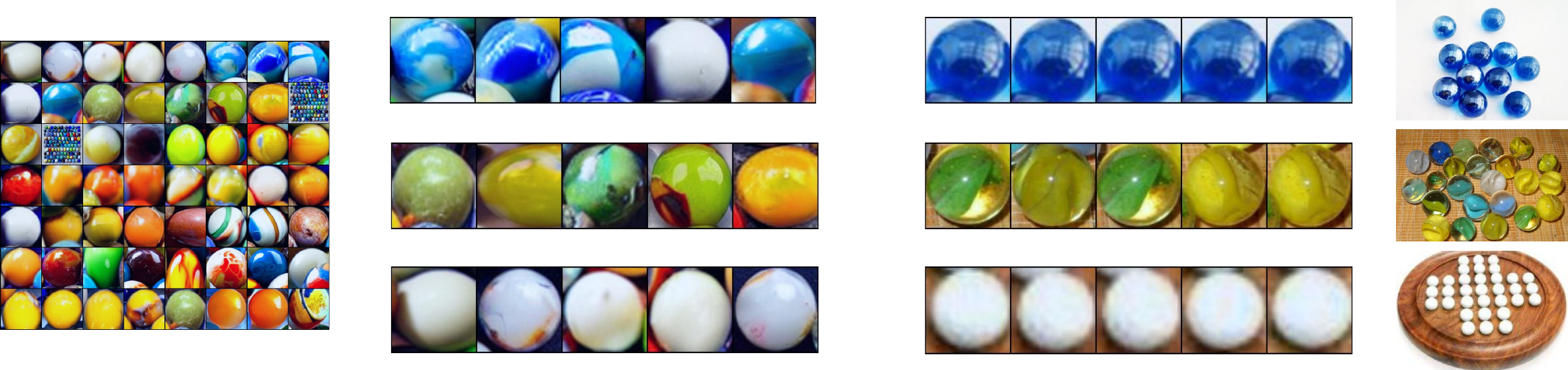}%
}
\caption{Visualization of the diffusion-generated patches for prototype generation and the corresponding selected patches from query images. Our method constructs prototypes according to the query images and selects the patches that most resemble the target objects.}
\label{fig: sd_prototype}
\end{center}
\end{figure*}

In this section, we provide additional qualitative analysis on prototype generation using Stable Diffusion. As shown in Figure \ref{fig: sd_prototype}, the patches generated by Stable Diffusion exhibit rich variations in terms of colors, shapes and textures. According to the query image, our method selects the patches that most resemble the query objects for constructing class prototypes. 

\section{Comparison with Using CLIP Features for Patch Selection}
\label{sec:comp_clip} 
To select class-relevant patches, we first construct a class prototype based on the corresponding textual description. Then we select the class-relevant patches among the object proposals using a nearest-neighbor lookup strategy. Another straightforward way is to directly apply this nearest-neighbor selection on the CLIP \cite{Radford2021LearningTV} features of the text and the object proposals. Here we compare this baseline method with ours. Specifically, we use the language encoder of a pre-trained CLIP (ViT-B/32) to extract text embeddings from the labels of the query classes, and its image encoder to extract visual embeddings for the query object proposals. We select the patches whose CLIP visual embeddings are the nearest neighbors of the corresponding CLIP's text embedding as class-relevant patches. Results are summerized in Table \ref{tab:comp_clip}. Using our generated prototypes for patch selection outperforms using CLIP features by a large margin. 

\vspace{6pt}
\begin{table*}[!h] 
  \centering
\resizebox{0.61\textwidth}{!}{%
  \begin{tabular}{c|cccc|cccc}
    \toprule
   \multirow{2}{*}{{\hspace{0.0cm}Class Prototypes}} & \multicolumn{4}{c|}{Val Set} & \multicolumn{4}{c}{Test Set} \\
    & MAE & RMSE & NAE & SRE & MAE & RMSE & NAE & SRE \\
    \midrule
   CLIP & 34.48 & 111.61 & 0.45 & 7.16 & 30.32 & 138.98 & 0.39 & 5.03 \\ 
   VAE-based & 27.47 & 90.85 & 0.37 & 4.52 & 23.14 & 114.40 & 0.34 & 3.95  \\ 
   SD-based & \textbf{26.30} & \textbf{88.80} & \textbf{0.34} & \textbf{4.27} & \textbf{21.53} & \textbf{113.28} & \textbf{0.31} & \textbf{3.61} \\ 
    \bottomrule
  \end{tabular}} \\ \vspace{6pt}
  \caption{ {Comparison with using CLIP features as prototypes to select class-relevant patches.} 
  }\label{tab:comp_clip}%
\end{table*}

\section{Multi-class Zero-Shot Counting}
\label{sec:multi_class}
\begin{figure*}[!ht]
\begin{center}
\hspace*{0.48cm}\includegraphics[width=0.96\columnwidth]{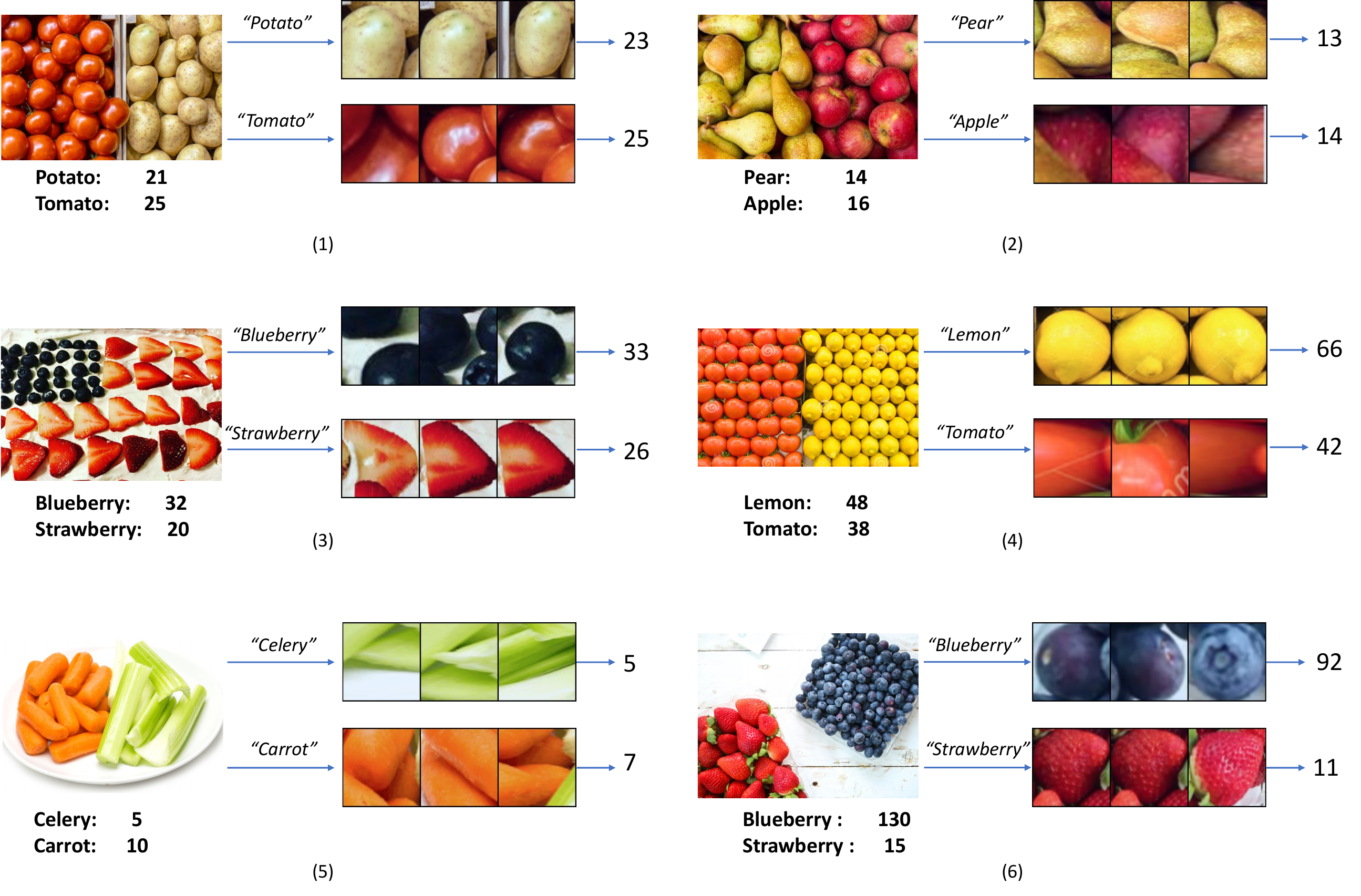}

\caption{ Visualizations of the selected patches: There are two classes with multiple object instances in a single image. To specify what to count, we provide the class name at test time. Our proposed method selects counting exemplars according to the given class name and count instances from the specific class.
} 
\label{fig:multi}
\end{center}

\end{figure*}

Figure \ref{fig:multi} provides additional visualizations of the selected patches in multi-class cases. 
As can be seen from the figure, our proposed method can select counting exemplars according to the given class name and count instances from that specific class in the input image. The reference-less methods i.e., RepRPN \cite{Ranjan2022Exemplar} and CounTR \cite{Liu2022CounTRTG}, can not be applied in this multi-class scenario.

\section{Ablation Study on the Number of Nearest Neighbors}
\label{sec:number_neighbor}
In the first step of our proposed patch selection method, we select the patches whose embeddings are the $N$-nearest neighbors of the class prototypes as class-relevant patches.
In this section, we provide an ablation study on the choice of $N$. We experiment with $5$, $10$, $15$ and $20$. The results are summarized in Table \ref{tab:num_k}. We observe that as the number of class-relevant patches (i.e., $N$) increases, the performance drops slightly. Setting $N$ too high might result in class-irrelevant patches being selected, which can decrease the performance. We observe the lowest MAE on the validation set when $N = 10$ and the lowest MAE on the test set when $N = 5$.

\begin{table*}[!h] 
  \centering
\resizebox{0.65\textwidth}{!}{%
  \begin{tabular}{c|cccc|cccc}
    \toprule
   \multirow{2}{*}{\parbox{2.2cm}{Number of Class-relevant Patches}} & \multicolumn{4}{c|}{Val Set} & \multicolumn{4}{c}{Test Set} \\
    & MAE & RMSE & NAE & SRE & MAE & RMSE & NAE & SRE \\
    \midrule
   5 & {26.80} & {91.87} & \textbf{0.34} & 4.32 & \textbf{21.53} & \textbf{113.28} & \textbf{0.31} & \textbf{3.61} \\ 
   10 & \textbf{26.30} & {88.80} & {0.34} & \textbf{4.27}& {21.81} & {124.92} & 0.31 & 3.71 \\ 
   15 & {26.31} & \textbf{88.32} & {0.34} & 4.33 & {21.91} & {126.72} & 0.31 & 3.73 \\ 
     20 & {26.37} & {88.50} & {0.34} & {4.31} & {22.13} & {126.76} & 0.32 & 3.76 \\
    \bottomrule
  \end{tabular}} \\ \vspace{6pt}
  \caption{ Ablation study on the number of class-relevant patches.
  }\label{tab:num_k}%

\end{table*}

\section{Failure Cases for SD-based Patch Selection}
\label{sec:failure}
\begin{figure*}[!ht]
\begin{center}
\hspace*{-0.7cm}\includegraphics[width=0.99\columnwidth]{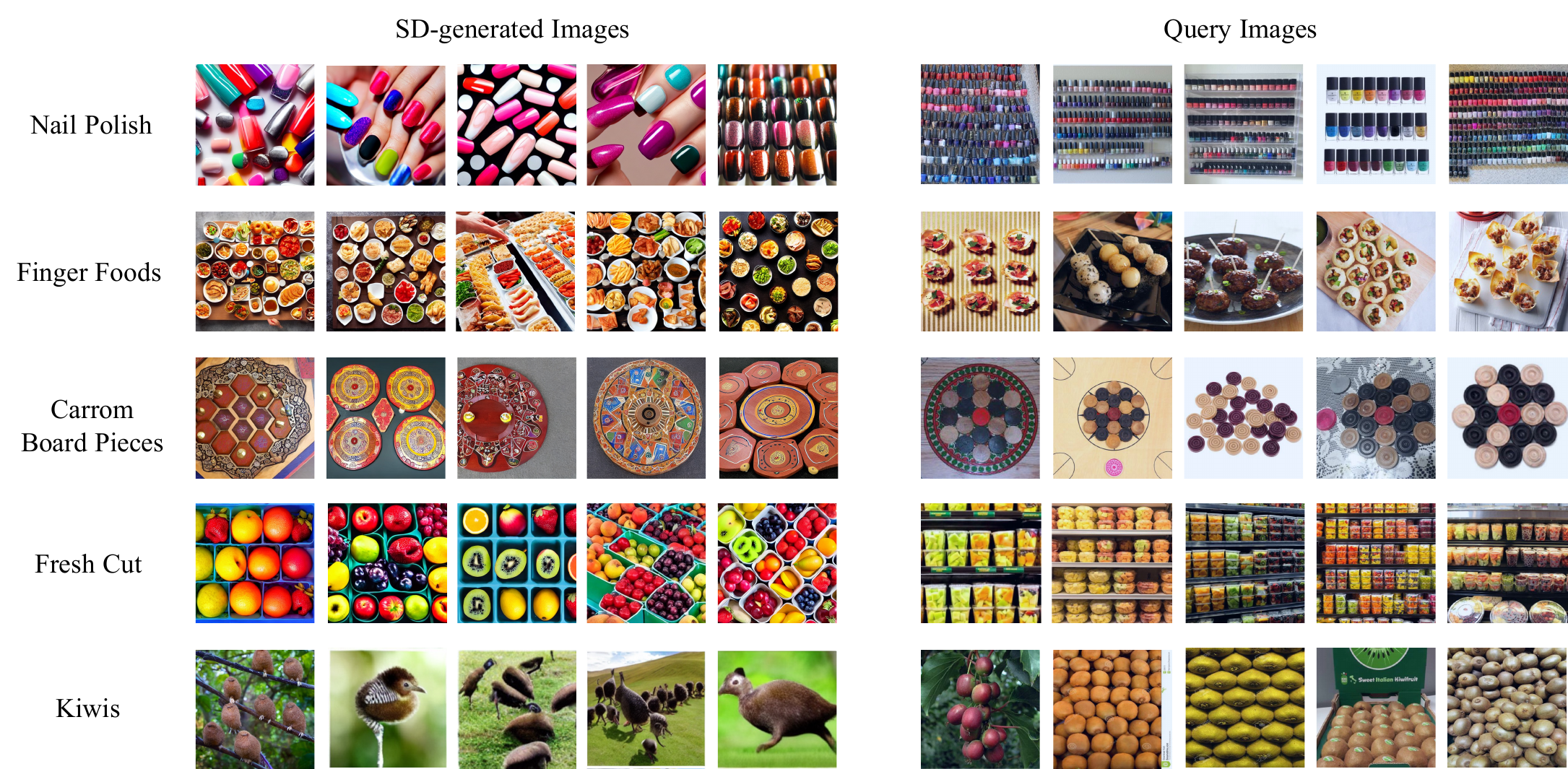}

\caption{ Visualizations of the SD-generated images and query images in some failure cases. Most cases are due to the ambiguity in the class labels which leads to mismatch between objects in the generated images and those present in the query images.
} 
\label{fig:failure}
\end{center}

\end{figure*}

In this section, we analyze the failure cases when employing the prototypes generated by Stable Diffusion (SD) for patch selection. A failure mode arises from the ambiguity in class labels.
For example, the term ``kiwi" could either refer to kiwi bird or kiwifruit. As shown in the last row of Figure \ref{fig:failure}, Stable Diffusion generates images of birds while the objects to count in the query images are kiwifruits. 
Another case where SD-generated prototypes would fail is when the class name is too generic to describe the objects accurately. For example, when provided with the class name ``fresh cut", Stable Diffusion generates images of different types of fruits, while the query images for FSC-147 \cite{Ranjan2021LearningTC} are cut fruits in plastic containers on shelves.
Another case is with a category labeled as ``nail polish". In this case, Stable Diffusion generates images of painted fingernails, while the query images are nail polish bottles on shelves. Although Stable Diffusion is capable of generating diverse and realistic images, it might generate images that do not resemble target images when provided with inaccurate / ambiguous prompts.
One way to resolve this issue is prompt engineering, i.e., designing and refining the prompts for more accurate image generation, which can be a future direction of this work. 

\section{Qualitative Analysis of Error Predictor}
\label{sec:qual_err} 
\begin{figure*}[!ht]
\begin{center}
\vspace*{-4cm}\includegraphics[width=0.6\columnwidth]{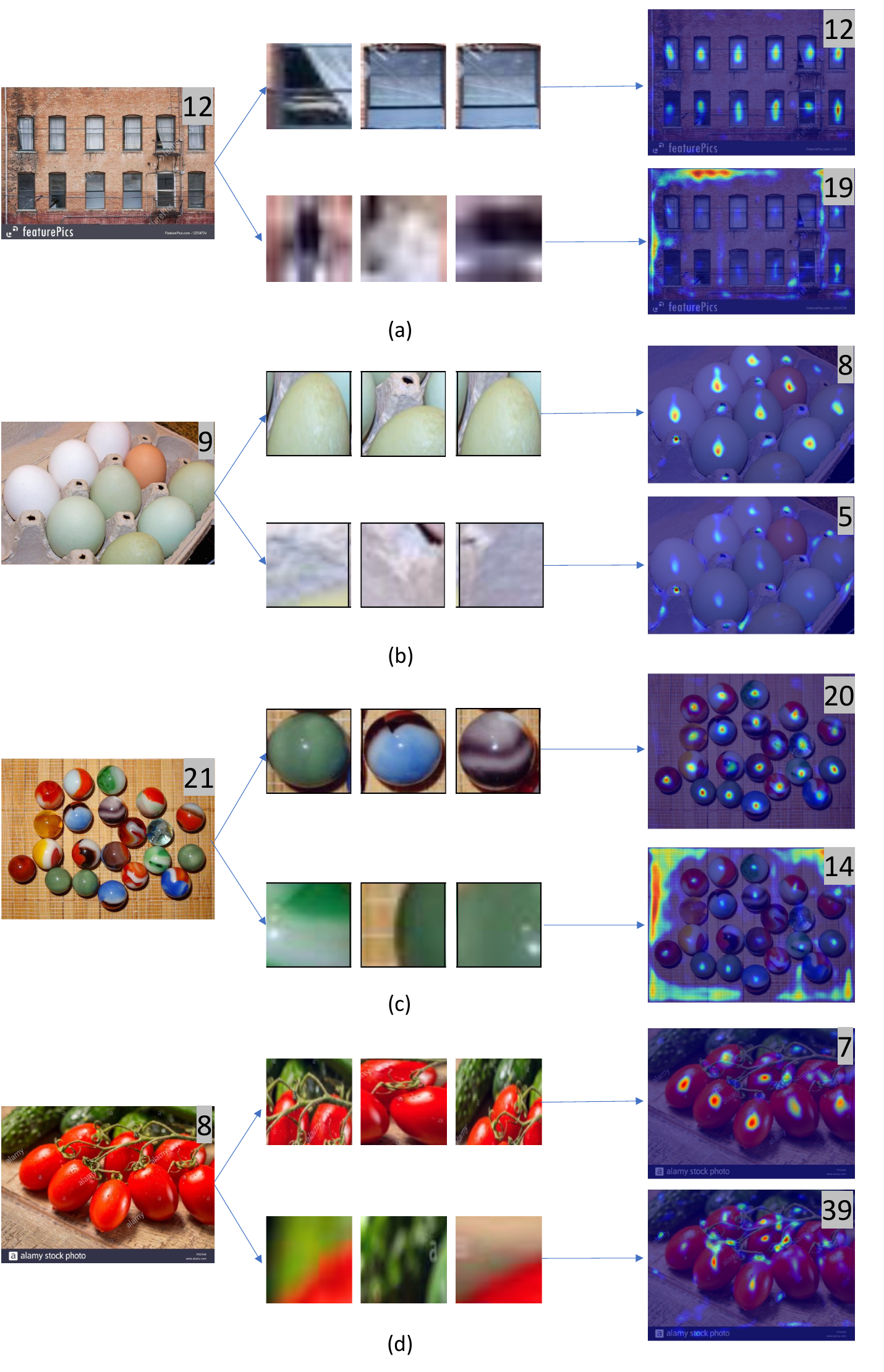}
\caption{ Visualizations of the patches with top-$3$ lowest and highest predicted counting errors. Our error predictor can pick out object crops results in accurate object counting.
} 
\label{fig:map}
\end{center}

\end{figure*}

In Figure \ref{fig:map}, we show a few input images and the corresponding patches with the  top-$3$ lowest and highest predicted counting errors. As can be seen from the figure, the patches with the smallest predicted errors are suitable to serve as counting exemplars and output meaningful density maps and accurate counting results. In comparison, the density maps produced by patches with the highest predicted errors fail to highlight the relevant image regions and lead to inaccurate counting results. This suggests that the predicted counting error can effectively indicate the goodness of the counting exemplars.

\bibliography{shortstrings,egbib}
\bibliographystyle{IEEEtran}